\documentclass[11pt]{article}
\usepackage{a4wide, chicago}
\usepackage{qtree}
\usepackage{graphicx}

\title{A Tutorial on the Expectation-Maximization Algorithm Including Maximum-Likelihood Estimation and EM Training of
Probabilistic Context-Free Grammars}
 
\author{%
  \begin{tabular}{c} 
    \textbf{Detlef Prescher} \\ Institute for Logic, Language and Computation \\ University of
    Amsterdam \\ {\tt prescher@science.uva.nl}
  \end{tabular}
}

\date{}

\newcommand{\comment}[1]{}

\newcommand{\pair}[2]
	   {({#1},{#2})}
\newcommand{\newcite}[1]
	   {\shortciteN{#1}}
\newtheorem{definition}
	   {Definition}
\newtheorem{axiom}
	   {Axiom}
\newtheorem{example}
	   {Example}
\newtheorem{theorem}
	   {Theorem}
\newcommand{\proof}
	   {\noindent\textbf{Proof\ }}

\newcommand{\treesize}[1]
	   {
	     \mbox{
	     \begin{scriptsize}
	       \hspace{-4ex}
	       \begin{tabular}{c} #1 \end{tabular}
	       \hspace{-4ex}
	     \end{scriptsize}
	     }
	   }
\newcommand{\nodesize}[1]
	   {\faketreewidth{\mbox{\hspace{#1}}}}
\newcommand{\append}[1]
	{\mbox{.}#1}
\newcommand{\rewr}{\ensuremath{\ \longrightarrow\ }}
		 
\newcommand{\lhs}{\ensuremath{\mbox{lhs}}}

\begin{document}

\maketitle

\section{Introduction}
The paper gives a brief review of the expectation-maximization
algorithm \cite{Dempster:77} in the comprehensible framework of
discrete mathematics. In Section~\ref{estimation}, two prominent
estimation methods, the relative-frequency estimation and the
maximum-likelihood estimation are presented. Section~\ref{em} is
dedicated to the expectation-maximization algorithm and a simpler
variant, the generalized expectation-maximization algorithm.  In
Section~\ref{dice}, two loaded dice are rolled. A more interesting
example is presented in Section~\ref{pcfgs}: The estimation of
probabilistic context-free grammars. Enjoy!

\section{Estimation Methods}
\label{estimation}
  
A statistics problem is a problem in which a
\textbf{corpus}\footnote{Statisticians use the term \textit{sample}
but computational linguists prefer the term \textit{corpus}} that has
been generated in accordance with some unknown \textbf{probability
distribution} must be analyzed and some type of inference about the
unknown distribution must be made.  In other words, in a statistics
problem there is a choice between two or more probability
distributions which might have generated the corpus. In practice,
there are often an infinite number of different possible distributions
-- statisticians bundle these into one single
\textbf{probability model} -- which might have generated the
corpus. By analyzing the corpus, an attempt is made to learn about the
unknown distribution. So, on the basis of the corpus, an
\textbf{estimation method} selects one \textbf{instance} of the
probability model, thereby aiming at finding the original distribution. In
this section, two common estimation methods, the relative-frequency
and the maximum-likelihood estimation, are presented.

\section*{Corpora}

\begin{definition}
\label{def.corpus} 
  Let $\mathcal{X}$ be a countable set. A real-valued function $f\!:
    \mathcal{X}\rightarrow\mathcal{R}$ is called a \textbf{corpus}, if
    $f$'s values are non-negative numbers \[f(x) \ge 0 \quad \mbox{
    for all } x \in \mathcal{X}\]
Each $x \in
    \mathcal{X}$ is called a \textbf{type}, and each value of f is
    called a \textbf{type frequency}. The 
\textbf{corpus size}\footnote{
Note that the corpus size $|f|$ is well-defined: The order of
  summation is not relevant for the value of the (possible infinite)
  series $\sum_{x \in \mathcal{X}}
  f(x)$, since the types
  are countable and the type frequencies are non-negative numbers}
   is defined as \begin{displaymath} |f| = \sum_{x \in \mathcal{X}}
  f(x) \end{displaymath} Finally, a corpus is called
  \textbf{non-empty} and
\textbf{finite} if \[0 < |f| < \infty\]
\end{definition}
In this definition, type frequencies are defined as
\textit{non-negative real numbers}.  The reason for not taking
\textit{natural numbers} is that some statistical estimation methods
define type frequencies as \textit{weighted} occurrence frequencies
(which are not natural but non-negative real numbers). Later on, in
the context of the EM algorithm, this point will become clear.  Note
also that a \textit{finite
corpus} might consist of an \textit{infinite number of types} with
positive frequencies.
The following definition shows that Definition~\ref{def.corpus}
covers the standard notion of the term \textit{corpus} (used in
Computational Linguistics) and of the term \textit{sample} (used in
Statistics).

\begin{definition}
Let $x_1,\ldots,x_n$ be a finite sequence of type instances from
$\mathcal{X}$. Each $x_i$ of this sequence is called a
\textbf{token}. The \textbf{occurrence frequency} of a type $x$ in the
sequence is defined as the following count
\begin{displaymath}
 f(x) = |\left\{\ i\ |\ x_i = x \right\}| 
\end{displaymath}
\end{definition}
Obviously, $f$ is a corpus in the sense of
Definition~\ref{def.corpus}, and it has the following properties: The
type $x$ does not occur in the sequence if $f(x)=0$; In any other case
there are $f(x)$ tokens in the sequence which are identical to
$x$. Moreover, the corpus size $|f|$ is identical to $n$, the number
of tokens in the sequence.

\section*{Relative-Frequency Estimation}

Let us first present the notion of probability that we use throughout
this paper.

\begin{definition}\label{def.prob_distrib}
Let $\mathcal{X}$ be a countable set of types. A real-valued function
$p\!:
\mathcal{X}\rightarrow\mathcal{R}$ is called a
\textbf{probability distribution on $\mathcal{X}$}, if $p$ has two properties: First, $p$'s values
are non-negative numbers \[p(x) \ge 0 \quad \mbox{ for all } x \in
\mathcal{X}\] and second, $p$'s values sum to 1
\begin{displaymath}
  \sum_{x \in \mathcal{X}} p(x) = 1
\end{displaymath}
\end{definition}
Readers familiar to probability theory will certainly note that we use
the term \textit{probability distribution} in a sloppy way
(\newcite{Duda:01}, page 611, introduce the term \textit{probability
mass function} instead). Standardly,
probability distributions allocate a probability value $p(A)$ to
subsets $A
\subseteq \mathcal{X}$, so-called \textbf{events} of an \textbf{event
space} $\mathcal{X}$, such that three specific axioms are satisfied
(see e.g. \newcite{DeGroot:89}):
\begin{axiom}
$p(A) \ge 0$ for any event $A$.
\end{axiom}
\begin{axiom}
$p(\mathcal{X})=1$.  
\end{axiom}
\begin{axiom}
\(
  p( \bigcup_{i=1}^\infty A_i ) =
  \sum_{i=1}^{\infty} p(A_i)
\)
for any infinite sequence of disjoint events $A_1, A_2, A_3,...$ 
\end{axiom}
Now, however, note that the probability distributions introduced in
Definition~\ref{def.prob_distrib} induce rather naturally the
following probabilities for events $A
\subseteq \mathcal{X}$
\[
  p(A) := \sum_{x \in A} p(x)
\]
Using the properties of $p(x)$, we can easily show that the
probabilities $p(A)$ satisfy the three axioms of probability
theory. So, Definition~\ref{def.prob_distrib} is justified and thus,
for the rest of the paper, we are allowed to put axiomatic probability
theory out of our minds.

\begin{definition}\label{def.relfreq}
Let $f$ be a non-empty and finite corpus. The probability distribution
\[\tilde{p}\!: \mathcal{X}\rightarrow\left[0,1\right]
\quad \mbox{ where } \quad \tilde{p}(x) = \frac{f(x)}{|f|}\] is called
the \textbf{relative-frequency estimate} on $f$.
\end{definition}
The relative-frequency estimation is the most comprehensible
estimation method and has some nice properties which will be discussed
in the context of the more general maximum-likelihood estimation.
For now, however, note that $\tilde{p}$ is well defined, since both
$|f|>0$ and $|f|< \infty$. Moreover, it is easy to check that
$\tilde{p}$'s values sum to one: $\sum_{x \in
\mathcal{X}} \tilde{p}(x) = \sum_{x \in \mathcal{X}} |f|^{-1} \cdot
f(x) = |f|^{-1} \cdot \sum_{x \in \mathcal{X}} f(x) = |f|^{-1} \cdot
|f| = 1$.

\section*{Maximum-Likelihood Estimation}

\begin{figure*}
\begin{center} 
\includegraphics[width=0.90\textwidth]{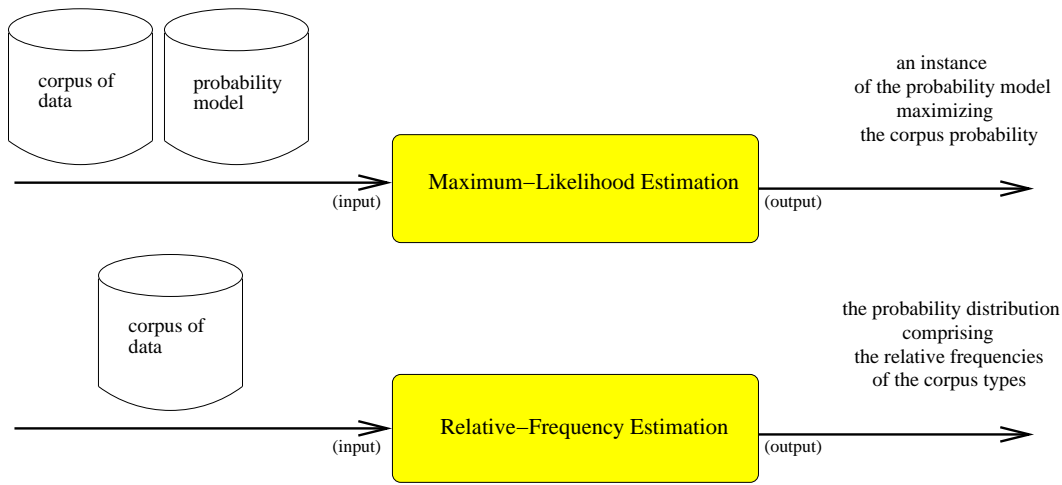}
\end{center}
\vspace{-2ex}
\caption{{\small Maximum-likelihood estimation and relative-frequency estimation}}
\label{figure.mle}
\end{figure*}

Maximum-likelihood estimation was introduced by R. A. Fisher in 1912,
and will typically yield an excellent estimate if the given corpus is
large. Most notably, maximum-likelihood estimators fulfill the
so-called \textbf{invariance principle} and, under certain conditions
which are typically satisfied in practical problems, they are even
\textbf{consistent estimators} \cite{DeGroot:89}.
For these reasons, maximum-likelihood estimation is probably the most
widely used estimation method. 

Now, unlike relative-frequency
estimation, maximum-likelihood estimation is a fully-fledged
estimation method that aims at selecting an instance of a
\textbf{given probability model} which might have originally generated
the given corpus. By contrast, the relative-frequency estimate is
defined on the basis of a corpus only (see
Definition~\ref{def.relfreq}). Figure~\ref{figure.mle} reveals the
conceptual difference of both estimation methods. In what follows, we
will pay some attention to describe the single setting, in which we
are exceptionally allowed to mix up both methods (see
Theorem~\ref{th.mle}). Let us start, however, by presenting the notion
of a probability model.
 
\begin{definition}\label{def.model}
A non-empty set $\mathcal{M}$ of probability distributions on a set
$\mathcal{X}$ of types is called a \textbf{probability model on
$\mathcal{X}$}. The elements of $\mathcal{M}$ are called
\textbf{instances of the model $\mathcal{M}$}. The
\textbf{unrestricted probability model} is the set
$\mathcal{M}(\mathcal{X})$ of \textit{all} probability distributions
on the set of types
\begin{displaymath}
  \mathcal{M}(\mathcal{X})
  =
  \left\{ 
  p\!: \mathcal{X}\rightarrow\left[0,1\right] 
  \ \left|
  \ \sum_{x \in \mathcal{X}} p(x) = 1
  \right.
  \right\}
\end{displaymath}
A probability model $\mathcal{M}$ is called
\textbf{restricted} in all other cases
\begin{displaymath}
   \mathcal{M} \subseteq \mathcal{M}(\mathcal{X})
   \quad\mbox{ and }\quad
   \mathcal{M} \not= \mathcal{M}(\mathcal{X})
\end{displaymath}
\end{definition}
In practice, most probability models are restricted since their
instances are often defined on a set $\mathcal{X}$ comprising
multi-dimensional types such that certain parts of the types are
statistically independent (see examples~\ref{ex.mle} and
\ref{em.dice}). Here is another side note: We already checked that
the relative-frequency estimate is a probability distribution, meaning
in terms of Definition~\ref{def.model} that \textit{the
relative-frequency estimate is an instance of the unrestricted
probability model}. So, from an extreme point of view, the
relative-frequency estimation might be also regarded as a
fully-fledged estimation method exploiting a corpus
\textbf{and} a probability model (namely, the unrestricted model).

In the following, we define maximum-likelihood estimation as a
method that aims at finding an instance of a given model which
maximizes the probability of a given corpus. Later on, we will see
that maximum-likelihood estimates have an additional property: They
are the instances of the given probability model that have a ``minimal
distance'' to the relative frequencies of the types in the corpus (see
Theorem~\ref{th.mle_and_entropy}). So, indeed, maximum-likelihood
estimates can be intuitively thought of in the intended way: They are
the instances of the probability model that might have originally
generated the corpus.

\begin{definition}\label{def.mle}
Let $f$ be a non-empty and finite corpus on a countable set
$\mathcal{X}$ of types. Let $\mathcal{M}$ be a probability model on
$\mathcal{X}$. The \textbf{probability of the corpus} allocated by an
instance $p$ of the model $\mathcal{M}$ is defined as
\begin{displaymath}
  L(f; p) = \prod_{x \in \mathcal{X}} p(x)^{f(x)}
\end{displaymath}
An instance $\hat{p}$ of the model $\mathcal{M}$ is called a
\textbf{maximum-likelihood estimate of $\mathcal{M}$ on $f$}, if and
only if the corpus $f$ is allocated a maximum probability by
$\hat{p}$
\begin{displaymath}
  L(f; \hat{p}) = \max_{p \in \mathcal{M}} L(f; p)
\end{displaymath}
(Based on continuity arguments, we use the convention that $p^0=1$ and $0^0=1$.)
\end{definition}
By looking at this definition, we recognize that maximum-likelihood
estimates are the solutions of a quite complex optimization
problem. So, some nasty questions about maximum-likelihood estimation
arise:
\begin{itemize}
\item[] \textbf{Existence} Is there for any probability model and any
  corpus a maximum-likelihood estimate of the model on the corpus?
\item[] \textbf{Uniqueness} Is there for any probability model and any
  corpus a unique maximum-likelihood estimate of the model on the
  corpus?
\item[] \textbf{Computability} For which probability models and
  corpora can maximum-likelihood
  estimates be efficiently computed?
\end{itemize}
\begin{figure*}
\begin{center}
\mbox{\includegraphics[width=0.50\textwidth]{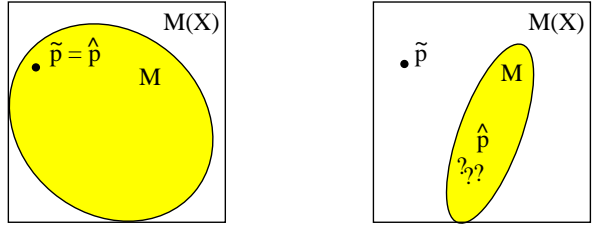}}
\vspace{-2ex}
\end{center}
\caption{{\small Maximum-likelihood estimation and relative-frequency
  estimation yield for some ``exceptional'' probability models the
  \textit{same estimate}. These models are lightly restricted or even
  unrestricted models that contain an instance comprising the relative
  frequencies of all corpus types (left-hand side). In practice,
  however, most probability models will not behave like that. So,
  maximum-likelihood estimation and relative-frequency estimation
  yield in most cases \textit{different estimates}. As a further and
  more serious consequence, the maximum-likelihood estimates have then
  to be searched for by genuine optimization procedures (right-hand
  side).}}
\vspace{+2ex}
\label{figure.mle.theorem}
\end{figure*}
For some probability models $\mathcal{M}$, the following theorem gives
a positive answer.

\begin{theorem}\label{th.mle}
Let $f$ be a non-empty and finite corpus on a countable set
$\mathcal{X}$ of types. Then:
\begin{itemize} 
\item[(i)] The relative-frequency estimate $\tilde{p}$ is a unique
  maximum-likelihood estimate of the unrestricted probability model
  $\mathcal{M}(\mathcal{X})$ on $f$.
\item[(ii)] The relative-frequency estimate $\tilde{p}$ is a
  maximum-likelihood estimate of a (restricted or unrestricted)
  probability model $\mathcal{M}$ on $f$, if and only if $\tilde{p}$
  is an instance of the model $\mathcal{M}$. In this case, $\tilde{p}$
  is a unique maximum-likelihood estimate of $\mathcal{M}$ on $f$.
\end{itemize}
\end{theorem}

\proof Ad~(i): Combine theorems~\ref{th.mle_and_entropy} and
  \ref{info.inequality}. Ad~(ii): ``$\Rightarrow$'' is
  trivial. ``$\Leftarrow$'' by (i)
\textbf{q.e.d.}
\\\\\noindent
On a first glance, proposition~(ii) seems to be more general than
proposition~(i), since proposition~(i) is about one single probability
model, the unrestricted model, whereas proposition~(ii) gives some
insight about the relation of the relative-frequency estimate to a
maximum-likelihood estimate of arbitrary restricted probability models
(see also Figure~\ref{figure.mle.theorem}). Both propositions,
however, are equivalent. As we will show later on, proposition (i) is
equivalent to the famous information inequality of information theory,
for which various proofs have been given in the literature. 

\begin{example}
On the basis of the following corpus
\begin{displaymath}
f(a) = 2,\ f(b)=3,\ f(c)=5
\end{displaymath}
we shall calculate the maximum-likelihood
estimate of the unrestricted probability model
\\$\mathcal{M}(\lbrace a,b,c\rbrace)$, as well as the maximum-likelihood estimate
of the restricted probability model 
\[
\mathcal{M}=\left\{ p
\in \mathcal{M}(\lbrace a,b,c\rbrace)\ \left. \begin{array}{c}\\\end{array}
\mbox{\hspace{-3ex}}\right| \ p(a)=0.5 
\right\}
\]
\end{example}
The solution is instructive, but is left to the reader.

\section*{The Information Inequality of Information Theory}

\begin{definition}\label{def.relative_entropy}
The \textbf{relative entropy} $D(p\ ||\ q)$ of the probability
distribution $p$ with respect to the probability distribution $q$ is
defined by
\begin{displaymath}
D(p\ ||\ q) = \sum_{x \in \mathcal{X}} p(x) \log \frac{p(x)}{q(x)}
\end{displaymath}
(Based on continuity arguments, we use the convention that $0
\log\frac{0}{q} = 0$ and $p \log\frac{p}{0} = \infty$ 
and $0 \log\frac{0}{0} = 0$. The logarithm
is calculated with respect to the base 2.)
\end{definition}
Connecting maximum-likelihood estimation with the concept of relative
entropy, the following theorem gives the important insight that the
relative-entropy of the relative-frequency estimate is minimal with
respect to a maximum-likelihood estimate.

\begin{theorem}\label{th.mle_and_entropy}
Let $\tilde{p}$ be the relative-frequency estimate on a non-empty and
finite corpus $f$, and let $\mathcal{M}$ be a probability model on the
set $\mathcal{X}$ of types. Then: An instance $\hat{p}$ of the model
$\mathcal{M}$ is a maximum-likelihood estimate of $\mathcal{M}$ on
$f$, if and only if the relative-entropy of $\tilde{p}$ is minimal with
respect to $\hat{p}$
\begin{displaymath}
  D(\tilde{p}\ ||\ \hat{p}) = \min_{p \in \mathcal{M}} D(\tilde{p}\
  ||\ p)
\end{displaymath}
\end{theorem}

\proof First,  the relative entropy
  $D(\tilde{p}\ ||\ p)$ is simply the difference of two further
  entropy values, the so-called \textbf{cross-entropy}
  $H(\tilde{p}; p) = - \sum_{x \in X} \tilde{p}(x) \log p(x)$ and the
  \textbf{entropy} $H(\tilde{p}) = - \sum_{x \in X} \tilde{p}(x) \log
  \tilde{p}(x)$ of the relative-frequency estimate
\begin{displaymath}
 D(\tilde{p}\ ||\ p) = H(\tilde{p}; p) - H(\tilde{p})
\end{displaymath}
(Based on continuity arguments and in full agreement with the
 convention used in Definition~\ref{def.relative_entropy}, we use here
 that $\tilde{p}\log 0 = - \infty$ and $0 \log0 = 0$.) It follows that
 minimizing the relative entropy is equivalent to minimizing the
 cross-entropy (as a function of the instances $p$ of the given
 probability model $\mathcal{M}$). The cross-entropy, however, is
 proportional to the negative log-probability of the corpus $f$
\begin{displaymath}
 H(\tilde{p}; p) = - \frac{1}{|f|} \log L(f; p)
\end{displaymath}
So, finally, minimizing the relative entropy $D(\tilde{p}\
||\ p)$ is equivalent to maximizing the corpus probability
$L(f; p)$.
\footnote{\label{foot.perp}For completeness, note that the
\textbf{perplexity} of a corpus $f$ allocated by a model instance $p$
is defined as $\mbox{perp}(f; p) = 2^{H(\tilde{p}; p)}$. This yields
$\mbox{perp}(f; p) = \sqrt[|f|]{\frac{1}{L(f; p)}}$ and $L(f; p) =
\left(\frac{1}{\mbox{perp}(f; p)}\right)^{|f|}$ as well as the 
common interpretation that \textbf{the perplexity value measures the
complexity of the given corpus from the model instance's view}: the
perplexity is equal to the size of an imaginary word list from which
the corpus can be generated by the model instance -- assuming that all
words on this list are equally probable. Moreover, the equations state that
minimizing the corpus perplexity $\mbox{perp}(f; p)$ is equivalent to
maximizing the corpus probability $L(f; p)$.}
\ \\
\ \\
\noindent
Together with Theorem~\ref{th.mle_and_entropy}, the following theorem, the so-called
information inequality of information theory, proves
Theorem~\ref{th.mle}. The information inequality states simply that
the relative entropy is a non-negative number -- which is zero, if and
only if the two probability distributions are equal.

\begin{theorem}[Information Inequality]\label{info.inequality}
Let $p$ and $q$ be two probability distributions. Then
\begin{displaymath}
 D(p\ ||\ q) \ge 0
\end{displaymath}
with equality if and only if $p(x) = q(x)$ for all $x \in \mathcal{X}$. 
\end{theorem} 

\proof See, e.g.,
\newcite{CoverThomas:91}, page 26.\\

\section*{*Maximum-Entropy Estimation}

Readers only interested in the expectation-maximization algorithm are
encouraged to omit this section. For completeness, however,
note that the relative entropy is
\textbf{asymmetric}. That means, in general
\begin{displaymath}
 D(p || q) \not= D(q || p)
\end{displaymath} 
It is easy to check that the triangle inequality is not valid too. So,
the relative entropy $D(.||.)$ is not a ``true'' distance function. On
the other hand, $D(.||.)$ has some of the properties of a distance
function. In particular, it is always non-negative and it is zero if
and only if $p=q$ (see Theorem~\ref{info.inequality}). So far,
however, we aimed at minimizing the relative entropy with respect to
its \textbf{second} argument, filling the first argument slot of
$D(.||.)$ with the relative-frequency estimate $\tilde{p}$. Obviously,
these observations raise the question, whether it is also possible to
derive other ``good'' estimates by minimizing the relative entropy
with respect to its \textbf{first} argument. So, in terms of
Theorem~\ref{th.mle_and_entropy}, it might be interesting to ask for
model instances $p^* \in
\mathcal{M}$ with
\begin{displaymath}
  D(p^* || \tilde{p}) = \min_{p \in \mathcal{M}} 
  D(p || \tilde{p})
\end{displaymath}
For at least two reasons, however, this initial approach of
relative-entropy estimation is too simplistic. First, it is tailored
to probability models that lack any generalization power. Second, it
does not provide deeper insight when estimating constrained
probability models. Here are the details:
\begin{itemize}
\item
   A closer look at Definition~\ref{def.relative_entropy} reveals that
   the relative entropy $D(p||\tilde{p})$ is finite for those model
   instances $p \in \mathcal{M}$ only that fulfill
\begin{displaymath} 
   \tilde{p}(x) = 0 \Rightarrow p(x) = 0
\end{displaymath}
   So, the initial approach would lead to model instances that are
   completely unable to generalize, since they are not allowed to
   allocate positive probabilities to at least some of the types not
   seen in the training corpus.
\item 
   Theorem~\ref{th.mle_and_entropy} guarantees that the
   relative-frequency estimate $\tilde{p}$ is a solution to the
   initial approach of relative-entropy estimation, whenever
   $\tilde{p} \in \mathcal{M}$. Now, Definition~\ref{def.MaxEnt}
   introduces the constrained probability models $M_{constr}$, and
   indeed, it is easy to check that $\tilde{p}$ is always an instance
   of these models.  In other words, estimating constrained
   probability models by the approach above does not result in
   interesting model instances.
\end{itemize}
Clearly, all the mentioned drawbacks are due to the fact that the
relative-entropy minimization is performed with respect to the
relative-frequency estimate. As a resource, we switch simply to a more
convenient reference distribution, thereby generalizing formally the
initial problem setting. So, as the final request, we ask for model
instances $p^*
\in \mathcal{M}$ with
\begin{displaymath}
  D(p^* || p_0) = \min_{p \in \mathcal{M}} D(p || p_0)
\end{displaymath}
In this setting, the \textbf{reference distribution} $p_0
\in \mathcal{M}(\mathcal{X})$ is a given instance of
the unrestricted probability model, and from what we have seen so far,
$p_0$ should allocate all types of interest a positive probability,
and moreover, $p_0$ should not be itself an instance of the
probability model $\mathcal{M}$.
\comment{Here are some candidates for a good reference distribution: 
(i)~$p_0$ = a PCFG tailored to the context-free backbone of a
unification grammar, and $p$ = a log-linear model. (ii)~$p_0$ the
uniform distribution on a large but finite subset $\mathcal{X}_0$ of
$\mathcal{X}$.}
Indeed, this request will lead us to the interesting maximum-entropy
estimates. Note first, that
\begin{displaymath}
  D(p || p_0) = H(p; p_0) - H(p)
\end{displaymath}
So, \textit{minimizing} $D(p || p_0)$ as a function of the model
instances $p$ is equivalent to \textit{minimizing} the cross entropy
$H(p; p_0)$ and \textit{simultaneously maximizing} the model entropy
$H(p)$. Now, simultaneous optimization is a hard task in general, and
this gives reason to focus firstly on maximizing the entropy $H(p)$ in
isolation. The following definition presents maximum-entropy
estimation in terms of the well-known maximum-entropy principle
\cite{Jaynes:57}. Sloppily formulated, the maximum-entropy principle
recommends to maximize the entropy $H(p)$ as a function of the
instances $p$ of certain ``constrained'' probability models.
\begin{definition}
\label{def.MaxEnt}
Let $f_1,\ldots,f_d$ be a finite number of real-valued functions on a
set $\mathcal{X}$ of types, the so-called \textbf{feature
functions}\footnote{Each of these feature functions can be thought of
as being constructed by inspecting the set of types, thereby measuring
a specific property of the types $x \in \mathcal{X}$. For example, if
working in a formal-grammar framework, then it might be worthy to look
(at least) at some feature functions $f_r$ directly associated to the
rules $r$ of the given formal grammar. The ``measure'' $f_r(x)$ of a
specific rule $r$ for the analyzes $x \in
\mathcal{X}$ of the grammar might be calculated, for example, in terms
of the occurrence frequency of $r$ in the sequence of those rules
which are necessary to produce $x$. For instance, \newcite{Chi:1999}
studied this approach for the context-free grammar formalism. Note,
however, that there is in general no recipe for constructing ``good''
feature functions: Often, it is really an intellectual challenge to
find those feature functions that describe the given data as best as
possible (or at least in a satisfying manner).}. Let $\tilde{p}$ be
the relative-frequency estimate on a non-empty and finite corpus $f$
on $\mathcal{X}$. Then, the
\textbf{probability model constrained by the expected values of 
$f_1\ldots f_d$ on $f$} is defined as
\begin{displaymath}
 \mathcal{M}_{constr} = \left\lbrace p \in \mathcal{M}(\mathcal{X}) \
 \left| \begin{array}{c}\\\\\end{array} E_pf_i = E_{\tilde{p}}f_i \
 \mbox{ for } i=1,\ldots, d \right\rbrace \right.
\end{displaymath}
Here, each $E_pf_i$ is the
\textbf{model
instance's expectation} of $f_i$
\begin{displaymath}
 E_{p}f_i = \sum_{x \in \mathcal{X}} p(x) f_i(x) 
\end{displaymath}
constrained to match $E_{\tilde{p}}f_i$, the \textbf{observed
 expectation} of
 $f_i$
\begin{displaymath}  
E_{\tilde{p}}f_i = \sum_{x \in \mathcal{X}}
\tilde{p}(x) f_i(x)
\end{displaymath}
Furthermore, a model instance $p^* \in \mathcal{M}_{constr}$ is called
a \textbf{maximum-entropy estimate of $\mathcal{M}_{constr}$} if and
only if
\begin{displaymath}
  H(p^*) = \max_{p \in \mathcal{M}_{constr}} H(p)
\end{displaymath}
\end{definition}
It is well-known that the maximum-entropy estimates have some nice
properties. For example, as Definition~\ref{def.ExponentialModels}
and Theorem~\ref{theorem.MaxEnt} show, they
can be identified to be the unique maximum-likelihood estimates of the
so-called exponential models (which are also known as log-linear
models).
\begin{definition}
\label{def.ExponentialModels}
Let $f_1,\ldots,f_d$ be a finite number of feature functions on a set
$\mathcal{X}$ of types. The \textbf{exponential model of
$f_1,\ldots,f_d$} is defined by
\begin{displaymath}
 \mathcal{M}_{exp} = \left\lbrace p \in \mathcal{M}(\mathcal{X}) \
 \left| \begin{array}{c}\\\\\end{array} p(x) = \frac{1}{Z_\lambda}
 e^{\lambda_1 f_1(x)+\ldots+\lambda_d f_d(x)} \mbox{ with }
 \lambda_1,\ldots,\lambda_d, Z_\lambda \in \mathcal{R}
\right\rbrace \right.
\end{displaymath}
Here, the \textbf{normalizing constant} $Z_\lambda$ (with $\lambda$ as
a short form for the sequence $\lambda_1,\ldots,\lambda_d$) guarantees
that $p \in
\mathcal{M}(\mathcal{X})$, and it is given by
\begin{displaymath}
Z_\lambda = \sum_{x \in \mathcal{X}} e^{\lambda_1
f_1(x)+\ldots+\lambda_d f_d(x)}
\end{displaymath}
\end{definition}
\begin{theorem}
\label{theorem.MaxEnt}
Let $f$ be a non-empty and finite corpus, and $f_1,\ldots,f_d$ be a
finite number of feature functions on a set $\mathcal{X}$ of
types. Then
\begin{itemize}
\item[(i)] 
   The maximum-entropy estimates of $\mathcal{M}_{constr}$ are
   instances of $\mathcal{M}_{exp}$, and the maximum-likelihood
   estimates of $\mathcal{M}_{exp}$ on $f$ are instances of
   $\mathcal{M}_{constr}$.
\item[(ii)] 
   If $p^* \in \mathcal{M}_{constr} \cap \mathcal{M}_{exp}$, then
   $p^*$ is both a unique maximum-entropy estimate of
   $\mathcal{M}_{constr}$ and a unique maximum-likelihood estimate of
   $\mathcal{M}_{exp}$ on $f$.
\end{itemize}
\end{theorem}
Part~(i) of the theorem simply suggests the form of the
maximum-entropy or maximum-likelihood estimates we are looking for. By
combining both findings of~(i), however, the search space is
drastically reduced for both estimation methods: We simply have to
look at the intersection of the involved probability models. In turn,
exactly this fact makes the second part of the theorem so valuable. If
there is a maximum-entropy \textbf{or} a maximum-likelihood estimate,
then it is in the intersection of both models, and thus according to
Part~(ii), it is a unique estimate, and even more, it is both a
maximum-entropy \textbf{and} a maximum-likelihood estimate.
\\\\
\proof See e.g. \newcite{CoverThomas:91}, pages 266-278.
For an interesting alternate proof of (ii), see
\newcite{Rat:97Report}. 
Note, however, that the proof of Ratnaparkhi's Theorem~1 is incorrect,
whenever the set $\mathcal{X}$ of types is infinite. Although
Ratnaparkhi's proof is very elegant, it relies on the existence of a
uniform distribution on $\mathcal{X}$ that simply does not exist in
this special case. By contrast, Cover and Thomas prove Theorem~11.1.1
without using a uniform distribution on $\mathcal{X}$, and so, they
achieve indeed the more general result.
\\\\\noindent
Finally, we are coming back to our request of minimizing the relative
entropy with respect to a given reference distribution $p_0 \in
\mathcal{M}(\mathcal{X})$. For constrained probability models,
the relevant results differ not much from the results described in
Theorem~\ref{theorem.MaxEnt}. So, let
\[
 \mathcal{M}_{ exp\cdot ref} = \left\lbrace p \in
 \mathcal{M}(\mathcal{X}) \ \left| \begin{array}{c}\\\\\end{array}
 p(x) = \frac{1}{Z_\lambda} e^{\lambda_1 f_1(x)+\ldots+\lambda_d
 f_d(x)} \cdot p_0(x) \mbox{ with } \lambda_1,\ldots,\lambda_d,
 Z_\lambda \in \mathcal{R}
\right\rbrace \right.
\]
Then, along the lines of the proof of Theorem~\ref{theorem.MaxEnt} it
can be also proven that the following propositions are valid.
\begin{itemize}
\item[(i)] 
   The minimum relative-entropy estimates of $\mathcal{M}_{constr}$ are
   instances of $\mathcal{M}_{exp\cdot ref}$, and the maximum-likelihood
   estimates of $\mathcal{M}_{exp\cdot ref}$ on $f$ are instances of
   $\mathcal{M}_{constr}$.
\item[(ii)] 
   If $p^* \in \mathcal{M}_{constr} \cap \mathcal{M}_{exp\cdot ref}$,
   then $p^*$ is both a unique minimum relative-entropy estimate of
   $\mathcal{M}_{constr}$ and a unique maximum-likelihood estimate of
   $\mathcal{M}_{exp\cdot ref}$ on $f$.
\end{itemize}
All results are displayed in Figure~\ref{figure.maxent.theorem}.
\begin{figure*}
\begin{center}
\begin{scriptsize}
\[
\mbox{\hspace{-3ex}}
\begin{array}{c}
 \mbox{\textbf{maximum-likelihood}}\\ \mbox{\textbf{estimation}}
\end{array}
\mbox{ of }
\left\{\mbox{\hspace{-2ex}}
 \begin{array}{c}
  \\
  \mbox{arbitrary probability models}
  \\\\\\
  \mbox{exponential models}
  \\\\\\
  \begin{array}{c}
   \mbox{exponential models}\\
   \mbox{with reference distributions}
  \\\\
  \end{array}
\end{array}
\mbox{\hspace{-2ex}}\right\}
\Longleftrightarrow
\left\{\mbox{\hspace{-2ex}}
\begin{array}{c}
   \\
   \begin{array}{c}
      \mbox{\textbf{minimum relative-entropy estimation}}\\
      \mbox{minimize $D(\tilde{p}||.)$}\\
   \mbox{($\tilde{p}$=relative-frequency estimate)}\\
   \end{array} 
   \\\\
   \begin{array}{c} 
      \mbox{\textbf{maximum-entropy estimation}}\\
      \mbox{\textbf{of constrained models}} 
   \end{array} 
   \\\\ 
   \begin{array}{c}
      \mbox{\textbf{minimum relative-entropy estimation}}\\
      \mbox{\textbf{of constrained models}}\\
      \mbox{minimize $D(.||p_0)$ }\\
            \mbox{($p_0$=reference distribution)}\\
   \\
   \end{array}
\end{array}
\mbox{\hspace{-2ex}}\right\}
\]
\end{scriptsize}
\end{center}
\vspace{-3ex}
\caption{{\small Maximum-likelihood estimation generalizes maximum-entropy
   estimation, as well as both variants of minimum relative-entropy
   estimation (where either the first or the second argument slot of
   $D(.||.)$ is filled by a given probability distribution).}}
\label{figure.maxent.theorem}
\end{figure*}

\section{The Expectation-Maximization Algorithm}
\label{em}

\begin{figure*}
\begin{center}
\mbox{\includegraphics[width=0.95\textwidth]{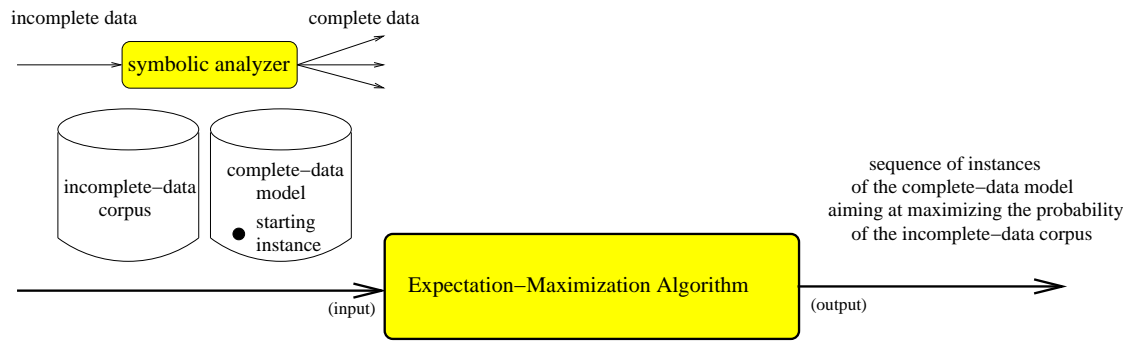}}
\end{center}
\vspace{-2ex}
\caption{{\small Input and output of the EM algorithm.}}
\label{figure.em.io}
\end{figure*}

The expectation-maximization algorithm was introduced by
\newcite{Dempster:77}, who also presented its main properties.  In
short, the EM algorithm aims at finding maximum-likelihood estimates
for settings where this appears to be difficult if not impossible. The
trick of the EM algorithm is to map the \textit{given data} to
\textit{complete data} on which it is well-known how to perform
maximum-likelihood estimation. Typically, the EM algorithm is applied
in the following setting:
\begin{itemize}
  \item Direct maximum-likelihood estimation of the given probability
    model on the given corpus is not feasible. For example, if the
    likelihood function is too complex (e.g. it is a product of sums).
  \item There is an obvious (but one-to-many) mapping to complete
    data, on which maximum-likelihood estimation can be easily
    done. The prototypical example is indeed that maximum-likelihood
    estimation on the complete data is already a solved problem.
\end{itemize}
Both relative-frequency and maximum-likelihood estimation are common
estimation methods with a two-fold input, a corpus and a probability
model\footnote{We associate the relative-frequency estimate with the
unrestricted probability model} such that the instances of the model
might have generated the corpus. The output of both estimation methods
is simply an instance of the probability model, ideally, the unknown
distribution that generated the corpus. In contrast to this setting,
in which we are almost \textit{completely informed} (the only thing
that is not known to us is the unknown distribution that generated the
corpus), the expectation-maximization algorithm is designed to
estimate an instance of the probability model for settings, in which
we are \textit{incompletely informed}. 

To be more specific, instead of a
\textbf{complete-data corpus}, the input of the
expectation-maximization algorithm is an
\textbf{incomplete-data corpus} together with a so-called \textbf{symbolic
analyzer}.  A symbolic analyzer is a device assigning to each
\textbf{incomplete-data type} a
\textbf{set of analyzes}, each analysis being a 
\textbf{complete-data type}. As a result, the missing complete-data corpus can
be partly compensated by the expectation-maximization algorithm: The
application of the the symbolic analyzer to the incomplete-data corpus
leads to an \textit{ambiguous complete-data corpus}. The ambiguity
arises as a consequence of the inherent \textbf{analytical ambiguity}
of the symbolic analyzer: the analyzer can replace each token of the
incomplete-data corpus by a set of complete-data types -- the set of
its analyzes -- but clearly, the symbolic analyzer is not able to
resolve the analytical ambiguity. 

The expectation-maximization algorithm performs a sequence of runs
over the resulting ambiguous complete-data corpus. Each of these runs
consists of an
\textbf{expectation step} followed by a \textbf{maximization step}. In
the \textbf{E step}, the expectation-maximization algorithm combines
the symbolic analyzer with an instance of the probability model. The
results of this combination is a \textbf{statistical analyzer} which
is able to
\textbf{resolve the analytical ambiguity} introduced by the symbolic
analyzer. In the \textbf{M step}, the expectation-maximization
algorithm calculates an ordinary maximum-likelihood estimate on the
resolved complete-data corpus. 

\begin{figure*}
\begin{center}
\mbox{\includegraphics[width=0.90\textwidth]{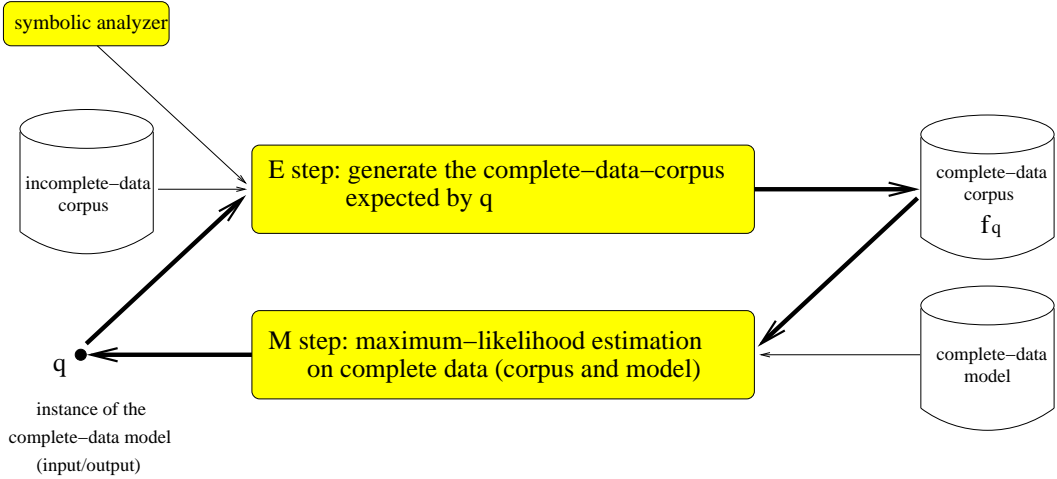}}
\end{center}
\caption{{\small Procedure of the EM algorithm. An incomplete-data
    corpus, a symbolic analyzer (a device assigning to each
    incomplete-data type a set of complete-data types), and a
    complete-data model are given. In the E step, the EM algorithm
    combines the symbolic analyzer with an instance $q$ of the
    probability model. The results of this combination is a
    statistical analyzer that is able to resolve the ambiguity of the
    given incomplete data. In fact, the statistical analyzer is used
    to generate an expected complete-data corpus $f_q$.  In the M
    step, the EM algorithm calculates an ordinary maximum-likelihood
    estimate of the complete-data model on the complete-data corpus
    generated in the E step. In further iterations, the estimates of
    the M-steps are used in the subsequent E-steps. The output of the
    EM algorithm are the estimates that are produced in the M steps.}}
\label{figure.em.procedure}
\end{figure*}

In general, however, a sequence of such runs is necessary. The reason
is that we never know which instance of the given probability model
leads to a good statistical analyzer, and thus, which instance of the
probability model shall be used in the E-step. The
expectation-maximization algorithm provides a simple but somehow
surprising solution to this serious problem. At the beginning, a
randomly generated
\textbf{starting instance} of the given probability model is used for the first
E-step. In further iterations, the estimate of the M-step is used for
the next E-step. Figure~\ref{figure.em.io} displays the input and the
output of the EM algorithm. The procedure of the EM algorithm is
displayed in Figure~\ref{figure.em.procedure}.

\section*{Symbolic and Statistical Analyzers}

\begin{definition}\label{def.symbolic_analyzer}
Let $\mathcal{X}$ and $\mathcal{Y}$ be non-empty and countable sets.
A function
\begin{displaymath}
     \mathcal{A}\!: \mathcal{Y} \rightarrow 2^\mathcal{X}
\end{displaymath}
is called a \textbf{symbolic analyzer} if the (possibly empty)
\textbf{sets of analyzes} $\mathcal{A}(y) \subseteq
\mathcal{X}$ are pair-wise disjoint, and 
the union of all sets of analyzes $\mathcal{A}(y)$ is complete
\begin{displaymath}
 \mathcal{X} = \sum_{y \in \mathcal{Y}} \mathcal{A}(y) 
\end{displaymath}
In this case, $\mathcal{Y}$ is called the set of
\textbf{incomplete-data types}, whereas $\mathcal{X}$ is called
the set of \textbf{complete-data types}. So, in other words, the
analyzes $\mathcal{A}(y)$ of the incomplete-data types $y$ form a
partition of the complete-data $\mathcal{X}$. Therefore, for each $x \in
\mathcal{X}$ exists a unique $y \in \mathcal{Y}$, the so-called
\textbf{yield} of $x$, such that
$x$ is an analysis of y
\[
   y=\mbox{yield}(x) \quad\mbox{if and only if}\quad 
   x \in \mathcal{A}(y)
\]
\end{definition}
For example, if working in a formal-grammar framework, the
\textit{grammatical sentences} can be interpreted as the
incomplete-data types, whereas the \textit{grammatical analyzes of the
sentences} are the complete-data types. So, in terms of
Definition~\ref{def.symbolic_analyzer}, a so-called \textit{parser} --
a device assigning a set of grammatical analyzes to a given sentence
-- is clearly a symbolic analyzer: The most important thing to check
is that the parser does not assign a given grammatical analysis to two
different sentences -- which is pretty obvious, if the
\textit{sentence words} are part of the grammatical analyzes.

\begin{definition}\label{def.statistical_analyzer}
A pair $<\!\mathcal{A},\ p\!>$ consisting of a symbolic analyzer
$\mathcal{A}$ and a probability distribution $p$ on the complete-data
types $\mathcal{X}$ is called a \textbf{statistical analyzer}. We use
a statistical analyzer to \textbf{induce probabilities} for the
incomplete-data types $y \in \mathcal{Y}$
\begin{displaymath}
  p(y) := \sum_{x \in \mathcal{A}(y)} p(x)  
\end{displaymath}  
Even more important, we use a statistical analyzer to
\textbf{resolve the analytical ambiguity} 
of an incomplete-data type $y \in \mathcal{Y}$ by looking at the
\textbf{conditional probabilities of the analyzes} $x \in \mathcal{A}(y)$
\begin{displaymath}
  p(x|y) := \frac{p(x)}{p(y)}
  \qquad\mbox{ where } y = \mbox{yield}(x)
\end{displaymath}
\end{definition}
It is easy to check that the statistical analyzer induces a proper
probability distribution on the set $\mathcal{Y}$ of incomplete-data
types
\begin{displaymath} 
    \sum_{y \in \mathcal{Y}} p(y) 
    =
    \sum_{y \in \mathcal{Y}} \sum_{x \in \mathcal{A}(y)} p(x)
    =
    \sum_{x \in \mathcal{X}} p(x)
    =
    1
\end{displaymath}
Moreover, the statistical analyzer induces also proper conditional
probability distributions on the sets of analyzes $\mathcal{A}(y)$
\begin{displaymath} 
    \sum_{x \in \mathcal{A}(y)} p(x|y) 
    =
    \sum_{x \in \mathcal{A}(y)} \frac{p(x)}{p(y)}
    =
    \frac{\sum_{x \in \mathcal{A}(y)} p(x)}{p(y)}
    =
    \frac{p(y)}{p(y)}
    =
    1
\end{displaymath}
Of course, by defining $p(x|y)=0$ for $y
\not=\mbox{yield}(x)$,
$p(.|y)$ is even a probability distribution on the full set
$\mathcal{X}$ of analyzes.

\section*{Input, Procedure, and Output of the EM Algorithm}

\begin{definition}\label{em.input}
The input of the expectation-maximization (EM) algorithm is
\begin{itemize}
\item[(i)]
 a \textbf{symbolic analyzer}, i.e., a function $\mathcal{A}$ which
 assigns a \textbf{set of analyzes} $\mathcal{A}(y) \subseteq
 \mathcal{X}$ to each \textbf{incomplete-data type} $y \in
 \mathcal{Y}$, such that all sets of analyzes form a partition of the
 set $\mathcal{X}$ of \textbf{complete-data types}
\begin{displaymath}
 \mathcal{X} = \sum_{y \in \mathcal{Y}} \mathcal{A}(y)
\end{displaymath} 
\item[(ii)] a non-empty and finite \textbf{incomplete-data corpus},
  i.e., a frequency distribution $f$ on the set of incomplete-data
  types
  \begin{displaymath}
    f\!: \mathcal{Y} \rightarrow \mathcal{R} \quad\mbox{ such that
  }\quad f(y) \ge 0 \mbox{ for all } y \in \mathcal{Y} \quad\mbox{
    and }\quad 0 < |f| < \infty
  \end{displaymath}
\item[(iii)] a \textbf{complete-data model $\mathcal{M} \subseteq
  \mathcal{M}(\mathcal{X})$}, i.e., each instance $p \in \mathcal{M}$
  is a probability distribution on the set of complete-data types
  \begin{displaymath}
    p\!: \mathcal{X} \rightarrow \left[0,1\right]
    \quad \mbox{ and } \quad \sum_{x \in \mathcal{X}} p(x) = 1
  \end{displaymath}
\item[(*)] \textbf{implicit input:} an \textbf{incomplete-data model
  $\mathcal{M} \subseteq \mathcal{M}(\mathcal{Y})$ } induced by the
  symbolic analyzer and the complete-data model. To see this, recall
  Definition~\ref{def.statistical_analyzer}. Together with a given
  instance of the complete-data model, the symbolic analyzer
  constitutes a statistical analyzer which, in turn, induces the following
  instance of the incomplete-data model  \begin{displaymath} p\!:
  \mathcal{Y} \rightarrow \left[0,1\right] \quad \mbox{ and } \quad
  p(y) = \sum_{x \in \mathcal{A}(y)} p(x) \end{displaymath} (Note: For
  both complete and incomplete data, the same notation symbols
  $\mathcal{M}$ and $p$ are used. The sloppy notation, however, is
  justified, because the incomplete-data model is a marginal of the
  complete-data model.)
\item[(iv)] a (randomly generated) \textbf{starting instance} $p_0$
  of the complete-data model $\mathcal{M}$.\\
  \noindent 
  (Note: If permitted by
  $\mathcal{M}$, then $p_0$ should not assign to any $x \in
  \mathcal{X}$ a probability of zero.)
\end{itemize}
\end{definition}

\begin{definition}\label{em.procedure}
The procedure of the EM algorithm is \\ \\
\noindent 
(1)\hspace{1.0cm} for each $i=1,\ 2,\ 3,\ ...$ do \ \\
(2)\hspace{1.5cm} $q := p_{i-1}$ \\ 
(3)\hspace{1.5cm} \textbf{E-step:} compute the \textbf{complete-data
  corpus} $f_q\!: \mathcal{X} \rightarrow \mathcal{R}$ \textbf{expected by $q$}
\begin{eqnarray*}
  &&\qquad
  f_q(x) \ := f(y) \cdot q(x|y) \qquad \mbox{ where } y = \mbox{yield}(x)
\end{eqnarray*}
(4)\hspace{1.5cm} \textbf{M-step:} compute a maximum-likelihood
   estimate $\hat{p}$ of
$\mathcal{M}$ on $f_q$ \ \\
\begin{displaymath}
 L(f_q; \hat{p}) = \max_{p \in \mathcal{M}} L(f_q, p)  
\end{displaymath}
\mbox{\hspace{3.7cm}(Implicit pre-condition of the EM algorithm: it
   exists!)}\ \\
(5)\hspace{1.5cm} $p_i := \hat{p}$
\ \\
(6)\hspace{1.0cm} end // for each $i$ \ \\
(7)\hspace{1.0cm} print $p_0, p_1, p_2, p_3, ...$ \
\end{definition}
\begin{figure*}
\begin{center}
\mbox{\includegraphics[width=0.80\textwidth]{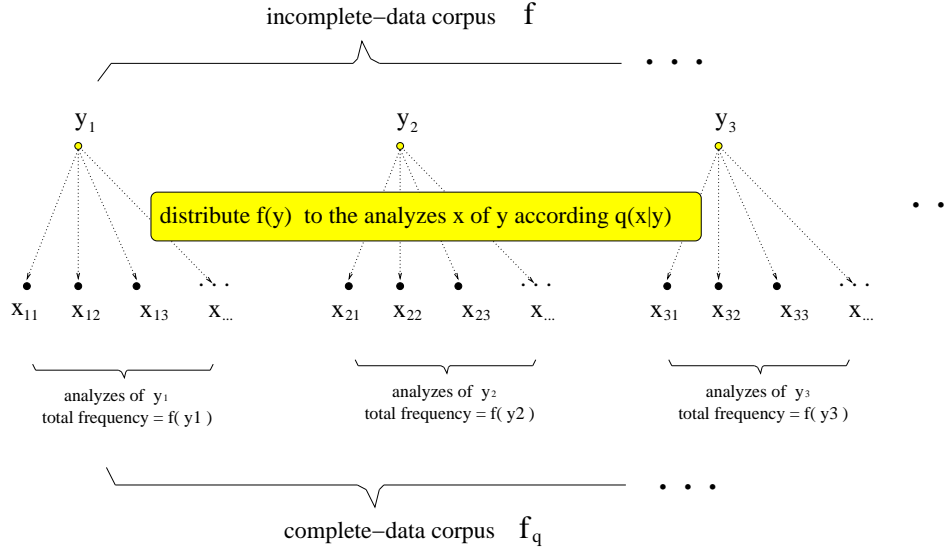}}
\vspace{-2ex}
\end{center}
\caption{{\small The E step of the EM algorithm. A complete-data
corpus $f_q(x)$ is generated on the basis of the incomplete-data
corpus $f(y)$ and the conditional probabilities $q(x|y)$ of the
analyzes of $y$. The frequency $f(y)$ is distributed among the
complete-data types $x \in \mathcal{A}(y)$ according to the
conditional probabilities $q(x|y)$. A simple reversed procedure
guarantees that the original incomplete-data corpus $f(y)$ can be
recovered from the generated corpus $f_q(x)$: Sum up all frequencies
$f_q(x)$ with $x \in \mathcal{A}(y)$. So the size of both corpora is the
same \( |f_q| = |f| \). \textit{Memory hook}: $f_q$ is the 
\textit{q}omplete data corpus.}}
\label{figure.em.e}
\end{figure*}
In line~(3) of the EM procedure, a complete-data corpus $f_q(x)$ has
to be generated on the basis of the incomplete-data corpus $f(y)$ and
the conditional probabilities $q(x|y)$ of the analyzes of $y$
(conditional probabilities are introduced in
Definition~\ref{def.statistical_analyzer}). In fact, this generation
procedure is conceptually very easy: according to the conditional
probabilities $q(x|y)$, the frequency $f(y)$ has to be distributed
among the complete-data types $x \in
\mathcal{A}(y)$. Figure~\ref{figure.em.e} displays the
procedure. Moreover, there exists a simple reversed procedure
(summation of all frequencies $f_q(x)$ with $x \in \mathcal{A}(y)$)
which guarantees that the original corpus $f(y)$ can be recovered from
the generated corpus $f_q(x)$. Finally, the size of both corpora is
the same
\begin{displaymath}
   |f_q| = |f|
\end{displaymath}
In line~(4) of the EM procedure, it is stated that a
maximum-likelihood estimate $\hat{p}$ of the complete-data model has
to be computed on the complete-data corpus $f_q$ expected by
$q$. Recall for this purpose that the probability of $f_q$ allocated
by an instance $p \in \mathcal{M}$ is defined as
\begin{displaymath}
  L(f_q; p) = \prod_{x \in \mathcal{X}} p(x)^{f_q(x)}  
\end{displaymath}
In contrast, the probability of the incomplete-data corpus $f$
allocated by an instance $p$ of the incomplete-data model is much more
complex. Using Definition~\ref{em.input}.*, we get an expression
involving a product of sums
\begin{displaymath}
  L(f; p) = 
  \prod_{y \in \mathcal{Y}} 
  \left(
  \sum_{x \in \mathcal{A}(y)} p(x)
  \right)^{f(y)}  
\end{displaymath}
Nevertheless, the following theorem reveals that the EM algorithm aims
at finding an instance of the incomplete-data model which possibly
maximizes the probability of the incomplete-data corpus.

\begin{theorem}\label{em.output}
The output of the EM algorithm is: A sequence of instances of the
complete-data model $\mathcal{M}$, the so-called \textbf{EM
re-estimates},
\begin{displaymath}
p_0,\ p_1,\ p_2,\ p_3,\ ...
\end{displaymath} 
such that the sequence of probabilities allocated to the
incomplete-data corpus is monotonic increasing
\begin{displaymath}
  L(f; p_0) \le L(f; p_1) \le L(f; p_2) \le L(f; p_3) \le \ldots
\end{displaymath}
\end{theorem}
It is common wisdom that the sequence of EM re-estimates will converge
to a (local) maximum-likelihood estimate of the incomplete-data model
on the incomplete-data corpus. As proven by \newcite{Wu:83}, however,
the EM algorithm will do this only in specific circumstances. Of
course, it is guaranteed that the sequence of corpus probabilities
(allocated by the EM re-estimates) must converge.  However, we are
more interested in the behavior of the EM re-estimates itself. Now,
intuitively, the EM algorithm might get stuck in a saddle point or
even a local minimum of the corpus-probability function, whereas the
associated model instances are hopping uncontrolled around (for
example, on a circle-like path in the ``space'' of all model
instances).
\\
\\
\proof See theorems~\ref{gem.output} and~\ref{em_and_gem}.

\section*{The Generalized Expectation-Maximization Algorithm}

The EM algorithm performs a sequence of maximum-likelihood estimations
on complete data, resulting in good re-estimates on incomplete-data
(``good'' in the sense of Theorem~\ref{em.output}). The following
theorem, however, reveals that the EM algorithm might overdo it
somehow, since there exist alternative M-steps which can be easier
performed, and which result in re-estimates having the same property
as the EM re-estimates.

\begin{definition}\label{gem.procedure}
A generalized expectation-maximization (GEM) algorithm has exactly
the same input as the EM-algorithm, but an easier M-step is performed
in its procedure: \\ \\
\noindent 
(4)\hspace{1.5cm} \textbf{M-step (GEM):} compute an instance $\hat{p}$ of
   the complete-data model $\mathcal{M}$ such that\\
   \begin{displaymath}
     L(f_q; \hat{p}) \ge L(f_q; q)
   \end{displaymath}
\end{definition}

\begin{theorem}\label{gem.output}
The output of a GEM algorithm is: A sequence of instances of the
complete-data model $\mathcal{M}$, the so-called \textbf{GEM
re-estimates}, such that the sequence of probabilities allocated to
the incomplete-data corpus is monotonic increasing.  
\end{theorem}

\proof 
Various proofs have been given in the literature. The first one was
presented by \newcite{Dempster:77}. For other variants of the EM
algorithm, the book of \newcite{McLachlan:97} is a good source.  Here,
we present something along the lines of the original proof.  Clearly,
a proof of the theorem requires somehow that we are able to express
\textit{the probability of the given incomplete-data corpus} 
$f$ in terms of the \textit{the probabilities of complete-data
corpora} $f_q$ which are involved in the M-steps of the GEM algorithm
(where both types of corpora are allocated a probability by the same
instance $p$ of the model $\mathcal{M}$). A certain entity, which we
would like to call the
\textbf{expected cross-entropy on the analyzes}, plays a major
role for solving this task. To be specific, the expected cross-entropy
on the analyzes is defined as the expectation of certain cross-entropy
values $H_{\mathcal{A}(y)}(q, p)$ which are calculated on the
different sets $\mathcal{A}(y)$ of analyzes. Then, of course, the
``expectation'' is calculated on the basis of the relative-frequency
estimate $\tilde{p}$ of the given incomplete-data corpus
\begin{displaymath}
   H_{\mathcal{A}}(q; p) = \sum_{y \in \mathcal{Y}} \tilde{p}(y) \cdot
   H_{\mathcal{A}(y)}(q;p)
\end{displaymath}
Now, for two instances $q$ and $p$ of the complete-data model, their
conditional probabilities $q(x|y)$ and $p(x|y)$ form proper
probability distributions on the set $\mathcal{A}(y)$ of analyzes of
$y$ (see Definition~\ref{def.statistical_analyzer}).  So, the
cross-entropy $H_{\mathcal{A}(y)}(q; p)$ on the set $\mathcal{A}(y)$
is simply given by
\begin{displaymath}
   H_{\mathcal{A}(y)}(q; p) = - \sum_{x \in \mathcal{A}(y)} q(x|y)
   \log p(x|y)
\end{displaymath}
Recalling the central task of this proof, a bunch of relatively
straight-forward calculations leads to the following interesting
equation
\footnote{
It is easier to show that
\[
H(\tilde{p}; p) = H(\tilde{p}_q; p) - H_{\mathcal{A}}(q; p).
\]
Here, $\tilde{p}$ is the relative-frequency estimate on the
incomplete-data corpus $f$, whereas $\tilde{p}_q$ is the
relative-frequency estimate on the complete-data corpus $f_q$.
However, by defining an ``average perplexity of the analyzes'',
$\mbox{perp}_{\mathcal{A}}(q;p) := 2^{H_{\mathcal{A}}(q; p)}$ (see
also Footnote~\ref{foot.perp}), the true spirit of the equation can be
revealed:
\[
   L(f_q;p) = 
   L(f;p) \cdot
   \left(
   \frac{1}{\mbox{perp}_{\mathcal{A}}(q; p)}
   \right)^{|f|}
\]
This equation states that the probability of a complete-data corpus
(generated by a statistical analyzer) is the product of the
probability of the given incomplete-data corpus and $|f|$-times the
average probability of the different corpora of analyzes (as generated
for each of the $|f|$ tokens of the incomplete-data corpus).}
\begin{displaymath}
   L(f;p) =
   \left(
   2^{H_{\mathcal{A}}(q; p)}
   \right)^{|f|} \cdot L(f_q;p)
\end{displaymath}
Using this equation, we can state that
\begin{displaymath}
   \frac{L(f;p)}{L(f;q)}
   =
   \left(
   2^{H_{\mathcal{A}}(q; p) - H_{\mathcal{A}}(q, q)}
   \right)^{|f|} \cdot \frac{L(f_q;p)}{L(f_q;q)}
\end{displaymath}
In what follows, we will show that, after each M-step of a GEM
algorithm (i.e. for $p$ being a GEM re-estimate $\hat{p}$),
both of the factors on the right-hand side of this equation are
not less than one.  First, an iterated application of the
information inequality of information theory (see
Theorem~\ref{info.inequality}) yields
\begin{eqnarray*}
   H_{\mathcal{A}}(q; p) - H_{\mathcal{A}}(q, q)
   &=&
   \sum_{y \in \mathcal{Y}} \tilde{p}(y) \cdot
   \left(
   H_{\mathcal{A}(y)}(q; p) - H_{\mathcal{A}(y)}(q; q) 
   \right)
   \\
   &=&
   \sum_{y \in \mathcal{Y}} \tilde{p}(y) \cdot
   D_{\mathcal{A}(y)}(q || p)
   \\\\
   &\ge& 
   0
\end{eqnarray*}
So, the first factor is never (i.e. for no model instance $p$) less than one
\begin{displaymath}
   \left(
   2^{H_{\mathcal{A}}(q; p) - H_{\mathcal{A}}(q, q)}
   \right)^{|f|} 
   \ge 1
\end{displaymath}
Second, by definition of the M-step of a GEM algorithm, the second
factor is also not less than one
\begin{displaymath}
  \frac{L(f_q;\hat{p})}{L(f_q;q)} \ge 1
\end{displaymath}
So, it follows
\begin{displaymath}
  \frac{L(f;\hat{p})}{L(f;q)} \ge 1
\end{displaymath}
yielding that the probability of the incomplete-data corpus allocated
by the GEM re-estimate $\hat{p}$ is not less than the probability of
the incomplete-data corpus allocated by the model instance $q$ (which
is either the starting instance $p_0$ of the GEM algorithm or the
previously calculated GEM re-estimate)
\begin{displaymath}
  L(f;\hat{p}) \ge L(f;q)
\end{displaymath}

\begin{theorem}\label{em_and_gem}
An EM algorithm is a GEM algorithm.  
\end{theorem} 

\proof In the M-step of an EM algorithm, a model instance $\hat{p}$ is
selected such that
\begin{displaymath}
 L(f_q; \hat{p}) = \max_{p \in \mathcal{M}} L(f_q, p)  
\end{displaymath}
So, especially 
\begin{displaymath}
 L(f_q; \hat{p}) \ge L(f_q, q)  
\end{displaymath}
and the requirements of the M-step of a GEM algorithm are met.

\section{Rolling Two Dice}
\label{dice}

\begin{example}\label{ex.dice}
We shall now consider an experiment in which two loaded dice are
rolled, and we shall compute the relative-frequency estimate on a
corpus of outcomes.
\end{example}
If we assume that the two dice are distinguishable, each outcome can
be represented as a pair of numbers $\pair{x_1}{x_2}$, where $x_1$ is
the number that appears on the first die and $x_2$ is the number that
appears on the second die. So, for this experiment, an appropriate set
$\mathcal{X}$ of types comprises the following 36 outcomes:
\begin{displaymath}
\begin{array}{c|cccccc}
(x_1,x_2) & x_2=1 & x_2=2 & x_2=3 & x_2=4 & x_2=5 & x_2=6\\ \hline
x_1=1 & \pair{1}{1} & \pair{1}{2} & \pair{1}{3} & \pair{1}{4} &
\pair{1}{5} & \pair{1}{6}\\ x_1=2 & \pair{2}{1} & \pair{2}{2} &
\pair{2}{3} & \pair{2}{4} & \pair{2}{5} & \pair{2}{6}\\ x_1=3 &
\pair{3}{1} & \pair{3}{2} & \pair{3}{3} & \pair{3}{4} & \pair{3}{5} &
\pair{3}{6}\\ x_1=4 & \pair{4}{1} & \pair{4}{2} & \pair{4}{3} &
\pair{4}{4} & \pair{4}{5} & \pair{4}{6}\\ x_1=5 & \pair{5}{1} &
\pair{5}{2} & \pair{5}{3} & \pair{5}{4} & \pair{5}{5} & \pair{5}{6}\\
x_1=6 & \pair{6}{1} & \pair{6}{2} & \pair{6}{3} & \pair{6}{4} &
\pair{6}{5} & \pair{6}{6}\\
\end{array}
\end{displaymath}
If we throw the two dice a 100~000 times, then the following
occurrence frequencies might arise
\begin{displaymath}
\begin{array}{c|cccccc}
f(x_1,x_2) & x_2=1 & x_2=2 & x_2=3 & x_2=4 & x_2=5 & x_2=6\\ \hline
 x_1=1 & 3790 & 3773 & 1520 & 1498 & 2233 & 2298 \\ x_1=2 & 3735 &
 3794 & 1497 & 1462 & 2269 & 2184 \\ x_1=3 & 4903 & 4956 & 1969 & 2035
 & 2883 & 3010 \\ x_1=4 & 2495 & 2519 & 1026 & 1049 & 1487 & 1451 \\
 x_1=5 & 3820 & 3735 & 1517 & 1498 & 2276 & 2191\\ x_1=6 & 6369 & 6290
 & 2600 & 2510 & 3685 & 3673
\end{array}
\end{displaymath}
The size of this corpus is $|f|=100~000$. So, the relative-frequency
estimate $\tilde{p}$ on $f$ can be easily computed (see
Definition~\ref{def.relfreq})

\begin{displaymath}
\begin{array}{c|cccccc}
\tilde{p}(x_1,x_2) & x_2=1 & x_2=2 & x_2=3 & x_2=4 & x_2=5 & x_2=6\\
 \hline x_1=1 & 0.03790 & 0.03773 & 0.01520 & 0.01498 & 0.02233 &
 0.02298 \\ x_1=2 & 0.03735 & 0.03794 & 0.01497 & 0.01462 & 0.02269 &
 0.02184 \\ x_1=3 & 0.04903 & 0.04956 & 0.01969 & 0.02035 & 0.02883 &
 0.03010 \\ x_1=4 & 0.02495 & 0.02519 & 0.01026 & 0.01049 & 0.01487 &
 0.01451 \\ x_1=5 & 0.03820 & 0.03735 & 0.01517 & 0.01498 & 0.02276 &
 0.02191\\ x_1=6 & 0.06369 & 0.06290 & 0.02600 & 0.02510 & 0.03685 &
 0.03673
\end{array}
\end{displaymath}

\begin{example}\label{ex.die}
We shall consider again Experiment~\ref{ex.dice} in which two loaded dice are
rolled, but we shall now compute the relative-frequency estimate on
the corpus of outcomes of the first die, as well as on the corpus of
outcomes of the second die.
\end{example}
If we look at the same corpus as in Example~\ref{ex.dice}, then the
corpus $f_1$ of outcomes of the first die can be calculated as
$f_1(x_1)=\sum_{x_2} f(x_1,x_2)$. An analog summation yields the
corpus of outcomes of the second die, $f_2(x_2)=\sum_{x_1}
f(x_1,x_2)$. Obviously, the sizes of all corpora are identical
$|f_1|=|f_2|=|f|=100~000$. So, the relative-frequency estimates
$\tilde{p_1}$ on $f_1$ and $\tilde{p_2}$ on $f_2$ are calculated as
follows
\begin{displaymath}
\begin{array}{c|c}
  f_1(x_1) & x_1 \\
\hline
  15112 &  1 \\
  14941 &  2 \\
  19756 &  3 \\
  10027 &  4 \\
  15037 &  5 \\
  25127 &  6 \\
\end{array}
\hspace{0.5cm}
\begin{array}{c|c}
  \tilde{p}_1(x_1) & x_1 \\
\hline
  0.15112 &  1 \\
  0.14941 &  2 \\
  0.19756 &  3 \\
  0.10027 &  4 \\
  0.15037 &  5 \\
  0.25127 &  6 \\
\end{array}
\hspace{2cm}
\begin{array}{c|c}
  f_2(x_2) & x_2 \\
\hline
  25112 &  1 \\
  25067 &  2 \\
  10129 &  3 \\
  10052 &  4 \\
  14833 &  5 \\
  14807 &  6 \\
\end{array}
\hspace{0.5cm}
\begin{array}{c|c}
  \tilde{p}_2(x_2) & x_2 \\
\hline
  0.25112 &  1 \\
  0.25067 &  2 \\
  0.10129 &  3 \\
  0.10052 &  4 \\
  0.14833 &  5 \\
  0.14807 &  6 \\
\end{array}
\end{displaymath}

\begin{example}\label{ex.mle}
  We shall consider again Experiment~\ref{ex.dice} in which two loaded
  dice are rolled, but we shall now compute a maximum-likelihood
  estimate of the probability model which assumes that the numbers
  appearing on the first and second die are statistically independent.
\end{example}
First, recall the definition of statistical independence 
(see e.g. \newcite{Duda:01}, page 613).
\begin{definition}
The variables $x_1$ and $x_2$ are said to be \textbf{statistically
independent} given a joint probability distribution $p$ on $\mathcal{X}$ if
and only if
\begin{displaymath}
  p(x_1,x_2) = p_1(x_1) \cdot p_2(x_2) 
\end{displaymath}
where $p_1$ and $p_2$ are the \textbf{marginal distributions} for $x_1$
and $x_2$
\begin{eqnarray*}
  p_1(x_1) = \sum_{x_2} p(x_1,x_2)\\  
  p_2(x_2) = \sum_{x_1} p(x_1,x_2)
\end{eqnarray*}
\end{definition}
So, let $\mathcal{M}_{1/2}$ be the probability model which assumes that the
  numbers appearing on the first and second die are statistically
  independent
\begin{displaymath}
  \mathcal{M}_{1/2}
  =
  \left\{
  p \in \mathcal{M}(\mathcal{X}) \ | \
  x_1 \mbox{ and } x_2 \mbox{ are statistically independent given } p 
  \right\}
\end{displaymath}
In Example~\ref{ex.dice}, we have calculated the relative-frequency
estimator $\tilde{p}$. Theorem~\ref{th.mle} states that $\tilde{p}$ is
the unique maximum-likelihood estimate of the unrestricted model
$\mathcal{M}(\mathcal{X})$. Thus, $\tilde{p}$ is also a candidate for
a maximum-likelihood estimate of $\mathcal{M}_{1/2}$.  Unfortunately,
however, $x_1$ and $x_2$ are \textbf{not} statistically independent given
$\tilde{p}$ (see e.g. $\tilde{p}(1,1) = 0.03790$ and
$\tilde{p}_1(1) \cdot \tilde{p}_2(1) = 0.0379\textbf{493}$). This has
two consequences for the experiment in which two (loaded) dice are
rolled:
\begin{itemize}
\item the probability model, which assumes that the numbers appearing
 on the first and second die are statistically independent, is a
 \textbf{restricted model} (see Definition~\ref{def.model}), and
\item \textbf{the relative-frequency estimate is in general not a
  maximum-likelihood estimate of the standard probability model}
  assuming that the numbers appearing on the first and second die are
  statistically independent.
\end{itemize}  
Therefore, we are now following Definition~\ref{def.mle} to compute
the maximum-likelihood estimate of $\mathcal{M}_{1/2}$.  Using the
independence property, the probability of the corpus $f$ allocated by
an instance $p$ of the model $\mathcal{M}_{1/2}$ can be calculated as
\begin{displaymath}
  L(f; p) 
  =
  \left(
  \prod_{ x_1 = 1,...,6} 
  p_1(x_1)^{f_1(x_1)}
  \right)
  \cdot
  \left(
  \prod_{ x_2 = 1,...,6} 
  p_2(x_2)^{f_2(x_2)} 
  \right)
  = 
  L(f_1; p_1) \cdot L(f_2;p_2)
\end{displaymath}
Definition~\ref{def.mle} states that the maximum-likelihood estimate
$\hat{p}$ of $\mathcal{M}_{1/2}$ on $f$ must maximize $L(f; p)$. A
product, however, is maximized, if and only if its factors are
simultaneously maximized. Theorem~\ref{th.mle} states that the corpus
probabilities $L(f_i; p_i)$ are maximized by the relative-frequency
estimators $\tilde{p_i}$. Therefore, the product of the
relative-frequency estimators $\tilde{p_1}$ and $\tilde{p_2}$ (on
$f_1$ and $f_2$ respectively) might be a candidate for the
maximum-likelihood estimate $\hat{p}$ we are looking for
\begin{displaymath}
  \hat{p}(x_1, x_2) = \tilde{p_1}(x_1) \cdot \tilde{p_2}(x_2)
\end{displaymath} 
Now, note that the marginal distributions of $\hat{p}$ are
identical with the relative-frequency estimators on $f_1$ and
$f_2$. For example, $\hat{p}$'s marginal distribution for 
$x_1$ is calculated as
\begin{displaymath}
  \hat{p}_1(x_1) = \sum_{x_2} \hat{p}(x_1,x_2)
  = \sum_{x_2} \tilde{p}_1(x_1) \cdot \tilde{p}_2(x_2)
  = \tilde{p_1}(x_1) \cdot \sum_{x_2} \tilde{p_2}(x_2)
  = \tilde{p}_1(x_1) \cdot 1 = \tilde{p}_1(x_1)
\end{displaymath}
A similar calculation yields $\hat{p}_2(x_2) = \tilde{p}_2(x_2)$.
Both equations state that $x_1$ and $x_2$ are indeed statistically
independent given $\hat{p}$
\begin{displaymath}
  \hat{p}(x_1, x_2) = \hat{p_1}(x_1) \cdot \hat{p_2}(x_2)
\end{displaymath}
So, finally, it is guaranteed that $\hat{p}$ is an instance of the
probability model $\mathcal{M}_{1/2}$ as required for a
maximum-likelihood estimate of $\mathcal{M}_{1/2}$. \textit{Note:
$\hat{p}$ is even an unique maximum-likelihood estimate since the
relative-frequency estimates $\tilde{p}_i$ are unique
maximum-likelihood estimates (see Theorem~\ref{th.mle}).} The
relative-frequency estimates $\tilde{p}_1$ and $\tilde{p}_2$ have
already been calculated in Example~\ref{ex.die}. So, $\hat{p}$ is
calculated as follows
\begin{displaymath}
\begin{array}{c|llllll}
\hat{p}(x_1,x_2) & x_2=1 & x_2=2 & x_2=3 & x_2=4 & x_2=5 & x_2=6\\
 \hline x_1=1 & 0.0379493 & 0.0378813 & 0.0153069 & 0.0151906 &
 0.0224156 & 0.0223763 \\x_1=2 & 0.0375198 & 0.0374526 & 0.0151337 &
 0.0150187 & 0.022162 & 0.0221231 \\x_1=3 & 0.0496113 & 0.0495224 &
 0.0200109 & 0.0198587 & 0.0293041 & 0.0292527 \\x_1=4 & 0.0251798 &
 0.0251347 & 0.0101563 & 0.0100791 & 0.014873 & 0.014847 \\x_1=5 &
 0.0377609 & 0.0376932 & 0.015231 & 0.0151152 & 0.0223044 & 0.0222653
 \\x_1=6 & 0.0630989 & 0.0629859 & 0.0254511 & 0.0252577 & 0.0372709 &
 0.0372055
\end{array}
\end{displaymath}

\begin{example}\label{em.dice}
We shall consider again Experiment~\ref{ex.dice} in which two loaded
dice are rolled. Now, however, we shall assume that we are
incompletely informed: the corpus of outcomes (which is given to us)
consists only of the sums of the numbers which appear on the first and
second die.  Nevertheless, we shall compute an estimate for a
probability model on the complete-data $(x_1,x_2) \in \mathcal{X}$.
\end{example}
If we assume that the corpus which is given to us was calculated on
the basis of the corpus given in Example~\ref{ex.dice}, then the
occurrence frequency of a sum $y$ can be calculated as $f(y) =
\sum_{x_1 + x_2 = y} f(x_1,x_2)$. These numbers are displayed in the
following table
\begin{center}
\begin{tabular}{r|r}
  $f(y)$ & $y$ \\
\hline
   3790 & 2 \\
   7508 &  3 \\
  10217 &  4 \\
  10446 &  5 \\
  12003 &  6 \\
  17732 &  7 \\
  13923 &  8 \\
   8595 &  9 \\
   6237 &  10 \\
   5876 &  11 \\
   3673 &  12 \\
\end{tabular}
\end{center}
For example, 
\begin{eqnarray*}
  f(4) 
  &=&
  f(1,3) + f(2,2) + f(3,1)
  = 
  1520 + 3794 + 4903
  =
  \underline{\underline{10217}}
\end{eqnarray*}
The problem is now, whether this corpus of sums can be used to
calculate a good estimate on the outcomes $(x_1,x_2)$ itself.
\textit{Hint: Examples~\ref{ex.dice} and~\ref{ex.mle} have shown that
a unique relative-frequency estimate $\tilde{p}(x_1,x_2)$ and a unique
maximum-likelihood estimate $\hat{p}(x_1,x_2)$ can be calculated on
the basis of the corpus $f(x_1,x_2)$. However, right now, this corpus
is not available!} Putting the example in the framework of the EM
algorithm (see Definition~\ref{em.input}), the set of
\textbf{incomplete-data types} is
\begin{displaymath}
  \mathcal{Y} = \left\{ 2, 3, 4, 5, 6, 7, 8, 9, 10, 11, 12 \right\}
\end{displaymath}
whereas the set of \textbf{complete-data types} is $\mathcal{X}$.
We also know the \textbf{set of analyzes for each incomplete-data type}
$y \in \mathcal{Y}$
\begin{displaymath}
  \mathcal{A}(y) = 
  \left\{ \pair{x_1}{x_2} \in \mathcal{X} \ | \ x_1 + x_2 = y \right\}
\end{displaymath}
As in Example~\ref{ex.mle}, we are especially interested in an
estimate of the (slightly restricted) \textbf{complete-data model}
$\mathcal{M}_{1/2}$ which assumes that the numbers appearing on the
first and second die are statistically independent.  So, for this
case, a randomly generated \textbf{starting instance} $p_0(x_1,x_2)$
of the complete-data model is simply the product of a randomly
generated probability distribution $p_{01}(x_1)$ for the numbers
appearing on the first dice, and a randomly generated probability
distribution $p_{02}(x_2)$ for the numbers appearing on the second dice
\begin{displaymath}
  p_0(x_1, x_2) = p_{01}(x_1) \cdot p_{02}(x_2)
\end{displaymath}
The following tables display some randomly generated numbers for
$p_{01}$ and $p_{02}$
\begin{displaymath}
\begin{array}{c|c}
  p_{01}(x_1) & x_1 \\ \hline 0.18 & 1 \\ 0.19 & 2 \\ 0.16 & 3 \\ 0.13 &
4 \\ 0.17 & 5 \\ 0.17 & 6 \\
\end{array}
\hspace{2cm}
\begin{array}{c|c}
  p_{02}(x_2) & x_2 \\ \hline 0.22 & 1 \\ 0.23 & 2 \\ 0.13 & 3 \\ 0.16 &
4 \\ 0.14 & 5 \\ 0.12 & 6 \\
\end{array}
\end{displaymath}
Using the random numbers for $p_{01}(x_1)$ and $p_{02}(x_2)$, a 
starting instance $p_0$ of the complete-data model $\mathcal{M}_{1/2}$
is calculated as follows
\begin{displaymath}
\begin{array}{c|cccccc}
p_0(x_1,x_2) & x_2=1 & x_2=2 & x_2=3 & x_2=4 & x_2=5 & x_2=6\\ \hline
 x_1=1 & 0.0396 & 0.0414 & 0.0234 & 0.0288 & 0.0252 & 0.0216 \\ x_1=2
 & 0.0418 & 0.0437 & 0.0247 & 0.0304 & 0.0266 & 0.0228 \\ x_1=3 &
 0.0352 & 0.0368 & 0.0208 & 0.0256 & 0.0224 & 0.0192 \\ x_1=4 & 0.0286
 & 0.0299 & 0.0169 & 0.0208 & 0.0182 & 0.0156 \\ x_1=5 & 0.0374 &
 0.0391 & 0.0221 & 0.0272 & 0.0238 & 0.0204 \\ x_1=6 & 0.0374 & 0.0391
 & 0.0221 & 0.0272 & 0.0238 & 0.0204 \\
\end{array}
\end{displaymath}
For example,
\begin{eqnarray*}
  p_0(1,3)
  &=&
  p_{01}(1) \cdot p_{02}(3)
  =
  0.18 \cdot 0.13
  =
  \underline{\underline{0.0234}}\\
  p_0(2,2)
  &=&
  p_{01}(2) \cdot p_{02}(2)
  =
  0.19 \cdot 0.23 
  =
  \underline{\underline{0.0437}}\\
  p_0(3,1)
  &=&
  p_{01}(3) \cdot p_{02}(1)
  =
  0.16 \cdot 0.22
  =
  \underline{\underline{0.0352}}\\
\end{eqnarray*}
So, we are ready to start the procedure of the EM algorithm.\\ \\
\noindent
\textbf{First EM iteration}. In the \textbf{E-step}, we shall compute
the \textbf{complete-data corpus} $f_q$ expected by $q := p_0$. For
this purpose, the probability of each incomplete-data type given the
starting instance $p_0$ of the complete-data model has to be computed
(see Definition~\ref{em.input}.*)
\begin{displaymath}
  p_0(y) = \sum_{x_1+x_2 = y} p_0(x_1,x_2)
\end{displaymath}
The above displayed numbers for $p_0(x_1,x_2)$ yield the following
instance of the incomplete-data model
\begin{displaymath}
\begin{array}{c|c}
  p_0(y) & y \\
\hline
0.0396 &   2 \\
0.0832 &   3 \\
0.1023 &   4 \\
0.1189 &  5 \\
0.1437 &  6 \\
0.1672 &  7 \\
0.1272 &  8 \\
0.0867 &  9 \\
0.0666 &  10 \\
0.0442 &  11 \\
0.0204 &  12 \\
\end{array}
\end{displaymath}
For example,
\begin{eqnarray*}
  p_0(4)
  &=&
  p_0(1,3) + p_0(2,2) + p_0(3,1)
  =
  0.0234 + 0.0437 + 0.0352
  =
  \underline{\underline{0.1023}}
\end{eqnarray*}
So, the complete-data corpus expected by $q := p_0$ is calculated as
follows (see line (3) of the EM procedure given in Definition~\ref{em.procedure})
\begin{displaymath}
\begin{array}{c|llllll}
f_q(x_1,x_2) & x_2=1 & x_2=2 & x_2=3 & x_2=4 & x_2=5 & x_2=6\\ \hline
 x_1=1 &
3790     &
3735.95  &
2337.03 & 
2530.23  &
2104.91  &
2290.74 \\
 x_1=2 &
3772.05  &
4364.45 & 
2170.03  &
2539.26 & 
2821     &
2495.63  \\
 x_1=3 & 
3515.53  &
3233.08  &
1737.39  &
2714.95 & 
2451.85 & 
1903.39 \\
 x_1=4 & 
2512.66  &
2497.49  &
1792.29  &
2276.72  &
1804.26  &
1460.92  \\
 x_1=5 & 
3123.95 & 
4146.66 & 
2419.01 & 
2696.47 & 
2228.84 & 
2712   \\
 x_1=6 &
3966.37 & 
4279.79 & 
2190.88 & 
2547.24 &
3164   & 
3673   
\end{array}
\end{displaymath}
For example,
\begin{eqnarray*}
  f_q(1,3) 
  &=&
  f(4) \cdot \frac{p_0(1,3)}{p_0(4)}
  =
  10217 \cdot \frac{0.0234}{0.1023}
  = 
  \underline{\underline{2337.03}}\\
  f_q(2,2) 
  &=&
  f(4) \cdot \frac{p_0(2,2)}{p_0(4)}
  =
  10217 \cdot \frac{0.0437}{0.1023}
  = 
  \underline{\underline{4364.45}}\\
  f_q(3,1) 
  &=&
  f(4) \cdot \frac{p_0(3,1)}{p_0(4)}
  =
  10217 \cdot \frac{0.0352}{0.1023}
  = 
  \underline{\underline{3515.53}}
\end{eqnarray*}
(The frequency $f(4)$ of the dice sum 4 is distributed to its analyzes 
(1,3),\ (2,2), and (3,1), simply by correlating the current
probabilities $q = p_0$ of the analyses...)
\\
\\
In the \textbf{M-step}, we shall compute a maximum-likelihood estimate
$p_1 := \hat{p}$ of the complete-data model $\mathcal{M}_{1/2}$ on the
complete-data corpus $f_q$. This can be done along the lines of
Examples~\ref{ex.die} and~\ref{ex.mle}.  \textit{Note: This is more or
less the trick of the EM-algorithm! If it appears to be difficult to
compute a maximum-likelihood estimate of an incomplete-data model then
the EM algorithm might solve your problem. It performs a sequence of
maximum-likelihood estimations on complete-data corpora. These corpora
contain in general more complex data, but nevertheless, it might be
well-known, how one has to deal with this data!} In detail: On the
basis of the complete-data corpus $f_q$ (where currently $q = p_0$), the corpus $f_{q 1}$ of
outcomes of the first die is calculated as $f_{q 1}(x_1)=\sum_{x_2}
f_q(x_1,x_2)$, whereas the corpus of outcomes of the second die is
calculated as $f_{q 2}(x_2)=\sum_{x_1} f_q(x_1,x_2)$. The following
tables display them:
\begin{displaymath}
\begin{array}{c|c}
  f_{q 1}(x_1) & x_1 \\
\hline
  16788.86 &  1 \\
  18162.42 &  2 \\
  15556.19 &  3 \\
  12344.34 &  4 \\
  17326.93 &  5 \\
  19821.28 &  6 \\
\end{array}
\hspace{2cm}
\begin{array}{c|c}
  f_{q 2}(x_2) & x_2 \\
\hline
  20680.56 &  1 \\
  22257.42 &  2 \\
  12646.63 &  3 \\
  15304.87 &  4 \\
  14574.86 &  5 \\
  14535.68 &  6 \\
\end{array}
\end{displaymath}
For example,
\begin{eqnarray*}
  f_{q 1}(1) 
  &=&
  f_q(1,1) + f_q(1,2) + f_q(1,3) + f_q(1,4) + f_q(1,5) + f_q(1,6)\\
  &=&
  3790 +
  3735.95 +
  2337.03 +
  2530.23 +
  2104.91 +
  2290.74 
  = \underline{\underline{16788.86}}
  \\
  f_{q 2}(1) 
  &=&
  f_q(1,1) + f_q(2,1) + f_q(3,1) + f_q(4,1) + f_q(5,1) + f_q(6,1)\\
  &=&
  3790 +
  3772.05 +
  3515.53 +
  2512.66 +
  3123.95 +
  3966.37
  = \underline{\underline{20680.56}} 
\end{eqnarray*}
The sizes of both corpora are still
$|f_{q 1}|=|f_{q 2}|=|f|=100~000$, resulting in the following relative-frequency
estimates ($p_{11}$ on $f_{q 1}$ respectively $p_{12}$ on $f_{q 2}$) 
\begin{displaymath}
\begin{array}{c|c}
  p_{11}(x_1) & x_1 \\ \hline 0.167889 & 1 \\ 0.181624 & 2 \\ 0.155562
  & 3 \\ 0.123443 & 4 \\ 0.173269 & 5 \\ 0.198213 & 6 \\
\end{array}
\hspace{2cm}
\begin{array}{c|c}
  p_{12}(x_2) & x_2 \\ \hline 0.206806 &1 \\ 0.222574 & 2 \\ 0.126466 & 3
  \\ 0.153049 & 4 \\ 0.145749 & 5 \\ 0.145357 & 6 \\
\end{array}
\end{displaymath}
So, the following instance is the
maximum-likelihood estimate of the model $\mathcal{M}_{1/2}$ on $f_q$
\begin{displaymath}
\begin{array}{c|llllll}
p_1(x_1,x_2) & x_2=1 & x_2=2 & x_2=3 & x_2=4 & x_2=5 & x_2=6\\ \hline
 x_1=1 & 0.0347204 & 0.0373677 & 0.0212322 & 0.0256952 & 0.0244696 &
 0.0244038 \\ x_1=2 & 0.0375609 & 0.0404247 & 0.0229692 & 0.0277973 &
 0.0264715 & 0.0264003 \\ x_1=3 & 0.0321711 & 0.034624 & 0.0196733 &
 0.0238086 & 0.022673 & 0.022612 \\ x_1=4 & 0.0255287 & 0.0274752 &
 0.0156113 & 0.0188928 & 0.0179917 & 0.0179433 \\ x_1=5 & 0.035833 &
 0.0385651 & 0.0219126 & 0.0265186 & 0.0252538 & 0.0251858 \\ x_1=6 &
 0.0409916 & 0.044117 & 0.0250672 & 0.0303363 & 0.0288893 &
 0.0288116\\
\end{array}
\end{displaymath}
For example,
\begin{eqnarray*}
  p_1(1,1) 
  &=&
  p_{11}(1) \cdot p_{12}(1) = 0.167889 \cdot 0.206806 
  = 
  \underline{\underline{0.0347204}}
  \\
  p_1(1,2) 
  &=&
  p_{11}(1) \cdot p_{12}(2) = 0.167889 \cdot 0.222574 
  = 
  \underline{\underline{0.0373677}}
  \\  
  p_1(2,1)
  &=&
  p_{11}(2) \cdot p_{12}(1) =  0.181624 \cdot 0.206806 
  = 
  \underline{\underline{0.0375609}}
  \\
  p_1(2,2) 
  &=&
  p_{11}(2) \cdot p_{12}(2) =  0.181624 \cdot 0.222574
  = 
  \underline{\underline{0.0404247}}
\end{eqnarray*}
So, we are ready for the second EM iteration, where an estimate $p_2$
is calculated. If we continue in this manner, we will arrive finally at the
\\ \\
\noindent
\textbf{1584th EM iteration}. The estimate which is calculated here is
\begin{displaymath}
\begin{array}{l|c}
  p_{1584, 1}(x_1) & x_1 \\ \hline
0.158396   &      1\\
0.141282    &     2\\
0.204291    &     3\\
0.0785532   &     4\\
0.172207    &     5\\
0.24527  & 6
\end{array}
\hspace{2cm}
\begin{array}{l|c}
  p_{1584,2}(x_2) & x_2 \\ \hline 
0.239281   &      1\\
0.260559   &      2\\
0.104026  &       3\\
0.111957  &       4\\
0.134419  &       5\\
0.149758  &       6
\end{array}
\end{displaymath}
yielding
\begin{displaymath}
\begin{array}{c|llllll}
p_{1584}(x_1,x_2) & x_2=1 & x_2=2 & x_2=3 & x_2=4 & x_2=5 & x_2=6\\ \hline
 x_1=1 &
0.0379012   & 
0.0412715   & 
0.0164773   & 
0.0177336   & 
0.0212914   & 
0.0237211  \\
 x_1=2 &
0.0338061   & 
0.0368123  &  
0.014697   &  
0.0158175  &  
0.018991   &  
0.0211581 \\
 x_1=3  &
0.048883   &  
0.0532299  &  
0.0212516  &  
0.0228718  &  
0.0274606  &  
0.0305942  \\
 x_1=4 &
0.0187963    &
0.0204678   & 
0.00817158   &
0.00879459  &
0.0105591   &
0.011764   \\
 x_1=5 &
0.0412059  &
0.0448701 & 
0.017914  & 
0.0192798  &
0.0231479  &
0.0257894 \\
 x_1=6 &
0.0586885  &
0.0639074  &
0.0255145  &
0.0274597  &
0.032969  & 
0.0367312   
\end{array}
\end{displaymath}
In this example, more EM iterations will result in exactly the same
re-estimates. So, this is a strong reason to quit the EM procedure.
Comparing $p_{1584,1}$ and $p_{1584,2}$ with the results of
Example~\ref{ex.die} \textit{(Hint: where we have assumed that a
complete-data corpus is given to us!)}, we see that the EM algorithm
yields pretty similar estimates.

\section{Probabilistic Context-Free Grammars}
\label{pcfgs}
This Section provides a more substantial example based on the
\textbf{context-free grammar} or \textbf{CFG} formalism, and it
is organized as follows: First, we will give some background
information about CFGs, thereby motivating that treating CFGs as
generators leads quite naturally to the notion of a probabilistic
context-free grammar (PCFG).  Second, we provide some additional
background information about ambiguity resolution by probabilistic
CFGs, thereby focusing on the fact that probabilistic CFGs can
resolve ambiguities, if the underlying CFG has a sufficiently high
expressive power. For other cases, we are pin-pointing to some useful
grammar-transformation techniques. Third, we will investigate the
standard probability model of CFGs, thereby proving that this model is
restricted in almost all cases of interest. Furthermore, we will give
a new formal proof that maximum-likelihood estimation of a CFG's
probability model on a corpus of trees is equal to the well-known and
especially simple treebank-training method. Finally, we will present
the EM algorithm for training a (manually written) CFG on a corpus of
sentences, thereby pin-pointing to the fact that EM training simply
consists of an iterative sequence of treebank-training steps. Small
toy examples will accompany all proofs that are given in this Section.

\section*{Background: Probabilistic Modeling of CFGs}

Being a bit sloppy (see e.g. \newcite{HopcroftUllman:79} for a formal
definition), a CFG simply consists of a finite set of
\textbf{rules}, where in turn, each rule consists of 
two parts being separated by a special symbol ``\rewr'', the so-called
\textbf{rewriting symbol}. The two parts of a rule are made up of
so-called
\textbf{terminal} and \textbf{non-terminal} symbols:
a rule's left-hand side simply consists of a single
non-terminal symbol, whereas the right-hand side 
is a finite sequence of terminal and non-terminal
symbols\footnote{As a consequence, the terminal and non-terminal
symbols of a given CFG form two finite and disjoint sets.}. Finally,
the set of non-terminal symbols contains at least one so-called
\textbf{starting symbol}. CFGs are also called
\textbf{phrase-structure grammars}, and the formalism is equivalent to
\textbf{Backus-Naur forms} or
\textbf{BNF} introduced by \newcite{Backus:1959}. In computational
linguistics, a CFG is usually used in two ways
\begin{itemize}
\item as a \textbf{generator}: a device for generating sentences, or 
\item as a \textbf{parser}: a device for assigning structure to a given sentence
\end{itemize}
In the following, we will briefly discuss these two issues. First of
all, note that in natural language, words do not occur in any
order. Instead, languages have constraints on word
order\footnote{Note, however, that so-called 
\textbf{free-word-order languages} (like Czech, German, or Russian) permit
many different ways of ordering the words in a sentence (without a
change in meaning). Instead of word order, these languages use case
markings to indicate who did what to whom.}. The central idea
underlying phrase-structure grammars is that words are organized into
\textbf{phrases}, i.e., grouping of words that form a unit.
Phrases can be detected, for example, by their ability (i) to stand
alone (e.g. as an answer of a wh-question), (ii) to occur in various
sentence positions, or by their ability (iii) to show uniform
syntactic possibilities for expansion or substitution. As an example,
here is the very first context-free grammar parse tree presented by
\newcite{Chomsky:1956}:

\vspace{-2ex}
\begin{center}
\treesize{
  \Tree
     [.Sentence [.NP the man ] [.VP [.Verb took ] [.NP the book ] ] ]
}
\end{center}

\noindent
As being displayed, Chomsky identified for the sentence ``the man took
the book'' (encoded in the leaf nodes of the parse tree) the following
phrases: two
\textbf{noun phrases}, ``the man'' and ``the book'' (the figure
displays them as NP subtrees), and one
\textbf{verb phrase}, ``took the book'' (displayed as VP subtree). The
following list of sentences, where these three phrases have been
substituted or expanded, bears some evidence for Chomsky's analysis:

\begin{scriptsize}
\begin{displaymath}
\left\{
\mbox{\begin{tabular}{c}
he\\ the man\\ the tall man\\ the very tall man\\ the tall man with sad eyes 
\end{tabular}}
\right\}
\mbox{took}
\left\{
\mbox{\begin{tabular}{c}
it\\ the book\\ the interesting book\\ the very interesting book\\
the very interesting book with 934 pages 
\end{tabular}}
\right\}
\end{displaymath}
\end{scriptsize}

\noindent
Chomsky's parse tree is based on the following CFG:

\begin{center}
\begin{tabular}{l}
Sentence \rewr NP VP\\
NP \rewr the man\\
NP \rewr the book\\
VP \rewr Verb NP\\
Verb \rewr took\\
\end{tabular}
\end{center}

\noindent
The CFG's terminal symbols are $\{$the, man, took, book$\}$, its
non-terminal symbols are $\{$Sentence, NP, VP, Verb$\}$, and its
starting symbol is ``Sentence''. Now, we are coming back to the
beginning of the section, where we mentioned that a CFG is usually
thought of in two ways: as a generator \textbf{or} as a parser. As a
generator, the example CFG might produce the following series of
intermediate parse trees (only the last one will be submitted to the
generator's output):\\

\vspace{-2ex}
\treesize{
\begin{tabular}{c}\mbox{\hspace{24ex}}\\Sentence\end{tabular} 
\Tree [.Sentence NP VP ]
\Tree [.Sentence [.NP the man ] VP ]
\Tree [.Sentence [.NP the man ] [.VP Verb NP ] ]
\\
\Tree [.Sentence [.NP the man ] [.VP [.Verb took ] NP ] ]
\Tree [.Sentence [.NP the man ] [.VP [.Verb took ] [.NP the book ] ] ] 
}

\noindent
Starting with the starting symbol, each of these intermediate parse
trees is generated by applying one rule of the CFG to a suitable
non-terminal leaf node of the previous parse tree, thereby adding the
CFG rule as a local tree. The generator stops, if all leaf nodes of
the current parse tree are terminal nodes. The whole generation
process, of course, is non-deterministic, and this fact will lead us
later on directly to probabilistic CFGs. As a parser, instead, the
example CFG has to deal with an input sentence like
\begin{center}
 ``the man took the book''
\end{center}

\noindent
Usually, the parser starts processing the input sentence by assigning
the words some local trees:

\vspace{-2ex}
\begin{center}
\treesize{\Tree [.NP the man ]} \treesize{\Tree [.Verb took ]}
\treesize{\Tree [.NP the book ]}
\end{center}

\noindent
Then, the parser tries to add more local trees, by processing all the
non-terminal nodes found in previous steps:

\vspace{-2ex}
\begin{center}
\treesize{\Tree [.NP the man ]} \treesize{\Tree [.VP [.Verb took ]
[.NP the book ] ]}
\end{center}

\noindent
Doing this recursively, the parser provides us with a parse tree of
the input sentence:

\vspace{-1ex}
\begin{center}
\treesize{\Tree [.Sentence [.NP the man ] [.VP
[.Verb took ] [.NP the book ] ] ]} 
\end{center}

\noindent
The example CFG is unambiguous for the given input sentence. Note,
however, that this is far away from being the common
situation. Usually, the parser stops, if all parse trees of the input
sentence have been generated (and submitted to the output).

Now, we demonstrate that the fact that we can understand CFGs as
generators leads directly to the
\textbf{probabilistic context-free grammar} or
\textbf{PCFG} formalism. As we already demonstrated for the generation
process, the rules of the CFG serve as local trees that are
incrementally used to build up a full parse tree (i.e. a parse tree
without any non-terminal leaf nodes). This process, however, is
non-deterministic: At most of its steps, some sort of \textbf{random
choice} is involved that selects one of the different CFG rules which
can potentially be appended to one of the non-terminal leaf nodes of
the current parse tree\footnote{Clearly, the final output of the
generator is directly affected by the specific rule that has been
selected by this random choice. Note also that there is another type
of uncertainty in the generation process, playing, however, only a
minor role: the specific place at which a CFG rule is to be appended
does obviously not affect the generator's final output. So, these
places can be deterministically chosen. For the generation process
displayed above, for example, we decided to append the local trees
always to the left-most non-terminal node of the actual partial-parse
tree.}. Here is an example in the context of the generation process
displayed above. For the CFG underlying Chomsky's very first parse
tree, the non-terminal symbol NP is the left-hand side of two rules:

\begin{center}
\begin{tabular}{l}
NP \rewr the man\\
NP \rewr the book\\
\end{tabular}
\end{center}

Clearly, when using the underlying CFG as a generator, we have to
select either the first or the second rule, whenever a local NP tree
shall be appended to the partial-parse tree given in the actual
generation step. The choice might be either
\textbf{fair} (both rules are chosen with probability $0.5$) or
\textbf{unfair} (the first rule is chosen, for example, with
probability $0.9$ and the second one with probability $0.1$). In
either case, a random choice between competing rules can be described
by probability values which are directly allocated to the rules:
\begin{displaymath}
0 \le p( \treesize{NP \rewr the man} ) \le 1
\quad\mbox{ and }\quad
0 \le p( \treesize{NP \rewr the book} ) \le 1
\end{displaymath}
such that
\begin{displaymath}
p( \treesize{NP \rewr the man} )\ + \ p( \treesize{NP \rewr the book}
) = 1
\end{displaymath}
Now, having these probabilities at hand, it turns out that it is even
possible to predict how often the generator will produce the one or
the other of the following alternate partial-parse trees:

\vspace{-3ex}
\begin{center}
\begin{tabular}{ll}
\treesize{\Tree [.Sentence [.NP the man ] [.VP [.Verb took ] NP ] ]} 
&
\treesize{\Tree [.Sentence [.NP the book ] [.VP [.Verb took ] NP ] ]} 
\\
\fbox{$p(\treesize{NP \rewr the man}) \cdot 100\%$}
&
\fbox{$p(\treesize{NP \rewr the book}) \cdot 100\%$}		
\end{tabular}
\end{center}
\vspace{+1ex}

\noindent
In turn, having this result at hand, we can also predict
how often the generator will produce full-parse trees, for example,
Chomsky's very first parse tree, or the parse tree of the sentence
``the book took the book'':

\begin{center}
\begin{tabular}{ll}
\treesize{\Tree [.Sentence [.NP the man ] [.VP [.Verb took ] [.NP the
book ] ] ]}
&
\treesize{\Tree [.Sentence [.NP the book ] [.VP [.Verb took ] [.NP the
book ] ] ]}
\\
\fbox{$p(\treesize{NP \rewr the man}) \cdot p(\treesize{NP \rewr the
book}) \cdot 100\%$}
&
\fbox{$p(\treesize{NP \rewr the book}) \cdot p(\treesize{NP \rewr the
book}) \cdot 100\%$}
\end{tabular}
\end{center}
\vspace{+1ex}

\noindent
So, if $p( \treesize{NP \rewr the man} ) = 0.9$ and $p( \treesize{NP
\rewr the book} ) = 0.1$, then it is nine times more likely that the
generator produces Chomsky's very first parse tree.
In the following, we are trying to generalize this result even a bit
more. As we saw, there are three rules in the CFG, which cause no
problems in terms of uncertainty. These are:
\begin{center}
\begin{tabular}{l}
Sentence \rewr NP VP\\
VP \rewr Verb NP\\
Verb \rewr took\\
\end{tabular}
\end{center}
To be more specific, we saw that these three rules have been always
deterministically added to the partial-parse trees of the generation
process. In terms of probability theory, determinism is expressed by
the fact that certain events occur with a probability of one. In other
words, a generator selects each of these rules with a probability of
$100\%$, either when starting the generation process, or when
expanding a VP or a Verb non-terminal node. So, we let
\begin{displaymath}
\begin{array}{l}
	p( \treesize{Sentence \rewr NP VP} ) = 1\\ 
	p( \treesize{VP \rewr Verb NP} ) = 1\\ 
	p( \treesize{Verb \rewr took} ) = 1
\end{array}
\end{displaymath}
The question is now: Have we won something by treating also the
deterministic choices as probabilistic events? The answer is yes: A
closer look at our example reveals that we can now predict easily how
often the generator will produce a specific parse tree: The likelihood
of a CFG's parse tree can be simply calculated as the product of the
probabilities of all rules occurring in the tree. For example:

\vspace{-3ex}
\begin{center}
\begin{tabular}{c}
\treesize{\Tree [.Sentence [.NP the man ] [.VP [.Verb took ] [.NP the
book ] ] ]} 
\\
\fbox{$p(\treesize{S \rewr NP VP}) \cdot p(\treesize{NP \rewr
the man}) \cdot p(\treesize{VP \rewr Verb NP}) \cdot p(\treesize{Verb
\rewr took}) \cdot p(\treesize{NP \rewr the
book})$}
\end{tabular}
\end{center}
\vspace{+1ex}

To wrap up, we investigated the small CFG underlying Chomsky's very
first parse tree. Motivated by the fact that a CFG can be used as
a generator, we assigned each of its rules a weight (a non-negative
real number) such that the weights of all rules with the same
left-hand side sum up to one. In other words, all CFG fragments
(comprising the CFG rules with the same left-hand side) have been
assigned a probability distribution, as displayed in the following
table:
\begin{center}
\begin{tabular}{ll}
\textbf{CFG rule}			& \textbf{Rule probability}\\
\hline
Sentence \rewr NP VP 	& $p(\treesize{Sentence \rewr NP VP}) = 1$\\
\begin{tabular}{l}
NP \rewr the man \\	
NP \rewr the book	
\end{tabular}
&
$\left. \begin{array}{l}
p(\treesize{NP \rewr the man})\\
p(\treesize{NP \rewr the book})
\end{array} \right\}$ \textbf{summing to 1}
\\
VP \rewr Verb NP & $p(\treesize{VP \rewr Verb NP}) = 1$\\ Verb \rewr
took & $p(\treesize{Verb \rewr took}) = 1$\\
\end{tabular}
\end{center}
As a result, the likelihood of each of the grammar's parse trees (when
using the CFG as a generator) can be calculated by multiplying the
probabilities of all rules occurring in the tree. This observation
leads directly to the standard definition of a probabilistic
context-free grammar, as well as to the definition of probabilities
for parse-trees.

\begin{definition}
\label{def.pcfg}
A pair $<G,p>$ consisting of a context-free grammar G and a
probability distribution $p\!:\mathcal{X}\rightarrow[0,1]$ on the set
$\mathcal{X}$ of all finite full-parse trees of $G$ is called a
\textbf{probabilistic context-free grammar} or
\textbf{PCFG}, if for all parse trees
$x \in \mathcal{X}$
\begin{displaymath}
	p(x) = \prod_{r \in G} p(r)^{f_r(x)}
\end{displaymath}
Here, $f_r(x)$ is the number of occurrences of the rule $r$ in the
tree $x$, and $p(r)$ is a probability allocated to the rule $r$, such
that for all non-terminal symbols $A$
\begin{displaymath}
	\sum_{r \in G_A} p(r) = 1
\end{displaymath}
where $G_A = \left\{ \ r \in G \ | \ \lhs(r) = A \right\}$ is the
grammar fragment comprising all rules with the left-hand side $A$.  In
other words, a probabilistic context-free grammar is defined by a
context-free grammar $G$ and some probability distributions on the
grammar fragments $G_A$, thereby inducing a probability distribution
on the set of all full-parse trees.
\end{definition}
So far, we have not checked for our example that the probabilities of
all full-parse trees are summing up to one. According to
Definition~\ref{def.pcfg}, however, this is
\textbf{the fundamental property} of PCFGs (and it should be really checked
for every PCFG which is accidentally given to us). Obviously, the
example grammar has four full-parse trees, and the sum of their
probabilities can be calculated as follows (by omitting all rules with
a probability of one):
\begin{eqnarray*}
p(\mathcal{X}) 
&=& 
p(\treesize{NP \rewr the man}) \cdot p(\treesize{NP \rewr the book})
+\\
&&
p(\treesize{NP \rewr the book}) \cdot p(\treesize{NP \rewr the book})
+\\
&&
p(\treesize{NP \rewr the man}) \cdot p(\treesize{NP \rewr the man})
+\\
&&
p(\treesize{NP \rewr the book}) \cdot p(\treesize{NP \rewr the man}) 
\\
&=&
\left( \begin{array}{c}\\\end{array}\mbox{\hspace{-2ex}}
p(\treesize{NP \rewr the man}) + p(\treesize{NP \rewr the book})
\right) \cdot p(\treesize{NP \rewr the book})
+\\
&&
\left( \begin{array}{c}\\\end{array}\mbox{\hspace{-2ex}}
p(\treesize{NP \rewr the man}) + p(\treesize{NP \rewr the book})
\right) \cdot p(\treesize{NP \rewr the man})
\\
&=& 1 
\end{eqnarray*}
For the last equation, we are using three times that $p$ is a
probability distribution on the grammar fragment $G_{\treesize{NP}}$,
i.e., we are exploiting that $p(\treesize{NP
\rewr the man}) + p(\treesize{NP \rewr the book})=1$.

The following examples show that we really have to do this kind of
``probabilistic grammar checking''. We are presenting two non-standard
PCFGs: The first one consists of the rules

\begin{center}
\begin{tabular}{ll}
	S \rewr NP sleeps 	& (1.0)\\
	S \rewr John sleeps	& (0.7)\\
	NP \rewr John           & (0.3)\\  
\end{tabular}
\end{center}

\noindent
The second one is a well-known highly-recursive grammar \cite{Chi:1998b}, and it
is given by

\vspace{-2ex}
\begin{center}
\begin{tabular}{lc}
	S \rewr S S 	& (q)\\
	S \rewr a	& (1-q)\\
\end{tabular}
with $0.5 < q \le 1$
\end{center}

\noindent
What is wrong with these grammars? Well, the first grammar provides us
with a probability distribution on its full-parse trees, as can be
seen here 

\vspace{-2ex}
\begin{center}
\begin{tabular}{ll}
\treesize{\Tree [.S [.NP John ] sleeps ]} 
&
\treesize{\Tree [.S John sleeps ]} 
\\
\fbox{$1.0 \cdot 0.3 = 0.3$}
&
\mbox{\hspace{5ex}}\fbox{$0.7$}		
\end{tabular}
\end{center}

\noindent
On each of its grammar fragments, however, the rule probabilities do not form a
probability distribution (neither on $G_{\treesize{S}}$ nor on
$G_{\treesize{NP}}$). The second grammar is even worse: We do have a
probability distribution on $G_{\treesize{S}}$, but even so, we do not
have a probability distribution on the set of full-parse trees (because their
probabilities are summing to less than one\footnote{
This can be proven
as follows: Let $\mathcal{T}$ be the set of all finite full-parse
trees that can be generated by the given context-free grammar. Then,
it is easy to verify that $\pi := p(\mathcal{T})$ is a solution of the
following equation
\[
	\pi = 1-q + q \cdot \pi \cdot \pi
\]
\vspace{-5ex}
\ \\
\noindent
Here, $1-q$ is the probability of the tree $\treesize{\Tree [.S a ]}\hspace{-7ex}$,
whereas $q
\cdot \pi \cdot \pi$ corresponds to the forest \treesize{\Tree [.S \mbox{\ $\mathcal{T}$\ } \mbox{\ $\mathcal{T}$\ } ]}\hspace{-10ex}~.
\\\noindent 
It is easy to check that
the derived quadratic equation has two solutions: $\pi_1=1$ and
$\pi_2=\frac{1-q}{q}$. Note that it is quite natural that \textbf{two}
solutions arise: The set of all ``infinite full-parse trees'' matches
also our under-specified approach of calculating $\pi$. Now, in the
case of $0.5 < q
\le 1$, it turns out that the set of infinite trees is allocated a
proper probability $\pi_1=1$. (For the special case $q=1$, this can be
intuitively verified: The generator will never touch the rule
\treesize{$S
\rewr a$}, and therefore, this special PCFG produces infinite parse
trees only.) As a consequence, all finite full-parse trees is allocated
the total probability $\pi_2$. In other words, $p(\mathcal{T}) =
\frac{1-q}{q} <
1$. In a certain sense, however, we are able to repair both
grammars. For example,
\begin{center}
\begin{tabular}{ll}
	S \rewr NP sleeps 	& (0.3)\\
	S \rewr John sleeps	& (0.7)\\
	NP \rewr John           & (1.0)\\  
\end{tabular}
\end{center}
is the standard-PCFG counterpart of the first grammar, where
\begin{center}
\begin{tabular}{lc}
	S \rewr S S 	& (1-q)\\
	S \rewr a	& (q)\\
\end{tabular}
with $0.5 < q \le 1$
\end{center}
is a standard-PCFG counterpart of the second grammar: The first
grammar and its counterpart provide us with exactly the same
parse-tree probabilities, while the second grammar and its counterpart
produce parse-tree probabilities, which are proportional to each
other. Especially for the second example, this interesting result is a
special case of an important general theorem recently proven by
\newcite{NederhofSatta:2003}. Sloppily formulated, their 
Theorem~7 states that:
\textit{For each \textbf{weighted CFG} (defined on the basis of \textbf{rule
weights} instead of rule probabilities) is a standard PCFG with the
same symbolic backbone, such that (i) the parse-tree probabilities
(produced by the PCFG) are summing to one, and (ii) the parse-tree
weights (produced by the weighted CFG) are proportional to the
parse-tree probabilities (produced by the PCFG).}
\\
As a consequence, we are getting what we really want: Applied
to ambiguity resolution, the original grammars and their counterparts
provide us with exactly the same maximum-probability-parse trees.
}).

\section*{Background: Resolving Ambiguities with PCFGs}

A property of most formalizations of natural language in terms of CFGs
is \textbf{ambiguity}: the fact that sentences have more than one
possible phrase structure (and therefore more than one meaning).  Here
are two prominent types of ambiguity:
\ \\\\
\begin{tabular}{l}
\hspace{-2ex}
\textit{Ambiguity caused by prepositional-phrase attachment:}
\\
\hspace{-2ex}
\treesize{    
\Tree [.S \qroof{Peter}.NP [.VP [.V saw ] \qroof{Mary}.NP \qroof{with
a telescope}.PP ] ]
	
\Tree [.S \qroof{Peter}.NP [.VP [.V saw ] [.NP \qroof{Mary}.NP \qroof{with
a telescope}.PP ] ] ] 
}		
\end{tabular}

\ \\\\
\begin{tabular}{l}
\hspace{-2ex}
\textit{Ambiguity caused by conjunctions}:
\\
\hspace{-2ex}
\treesize{    
\hspace{-6ex}
\Tree [.S [.NP  \qroof{the mother}.NP [.PP [.P of ] [.NP \qroof{the
boy}.NP [.CONJ and ] \qroof{the girl}.NP ] !\nodesize{24ex} ] ]
!\nodesize{70ex} \qroof{left}.VP ]
\hspace{-15ex}
\Tree [.S [.NP [.NP  \qroof{the mother}.NP [.PP [.P of ] \qroof{the
boy}.NP ] ]
!\nodesize{20ex} [.CONJ and ] \qroof{the girl}.NP ] \qroof{left}.VP ]
}
\end{tabular}

\ \\\\
\noindent
As usual in computational linguistics, some phrase structures have
been displayed in abbreviated form: For example, the term \treesize{
\qroof{the mother}.NP } is used as a short form for the parse tree
\treesize{ \Tree [.NP [.DET the ] [.N mother ] ] }, and
the term \treesize{\qroof{of the boy}.PP } is a place holder for the
even more complex parse tree \treesize{ \Tree [.PP [.P of ] [.NP
[.DET the ] [.N boy ] ] ] }.
\ \\\\
In both examples, the ambiguity is caused by the fact that the
underlying CFG contains
\textbf{recursive rules}, i.e., rules that can be applied an arbitrary
number of times. Clearly, the rules
\mbox{NP \rewr NP CONJ NP} and \mbox{NP \rewr NP PP} belong to this
type, since they can be used to generate nominal phrases of an
arbitrary length. The rules \mbox{VP \rewr V NP} and \mbox{PP \rewr P
NP}, however, might be also called (indirectly) recursive, since they
can generate verbal and prepositional phrases of an arbitrary length
(in combination with \mbox{NP \rewr NP PP}). Besides ambiguity,
recursivity makes it also possible that two words that are generated
by the same CFG rule (i.e. which are syntactically linked) can occur
far apart in a sentence:
\begin{quote}
   The \textbf{bird} with the nice brown eyes and the beautiful tail
   feathers \textbf{catches} a worm.
\end{quote}
These types of phenomena are called \textbf{non-local dependencies},
and it is important to note that non-local phenomena (which can be
handled by CFGs) are beyond the scope of many popular models that
focus on modeling local dependencies (such as n-gram, Markov, and
hidden Markov models\footnote{It is well-known, however, that a CFG
without \textbf{center-embeddings} can be transformed to a regular
grammar (the symbolic backbone of a hidden Markov model).}).  So, a
part-of-speech tagger (based on a HMM model) might have difficulties
with sentences like the one we mentioned, because it will not expect
that a singular verb occurs after a plural noun.

Having this at hand, of course, the central question is now: Can PCFGs
handle ambiguity?  The somewhat surprising answer is: Yes, but the
symbolic backbone of the PCFG plays a major role in solving this
difficult task. To be a bit more specific, the CFG underlying the
given PCFG has to have some good properties, or the other way round,
probabilistic modeling of some ``weak'' CFGs may result in PCFGs which
can not resolve the CFG's ambiguities. From a probabilistic modeler's
point of view, there is really some non-trivial relation between such
tasks as ``writing a formal grammar'' and ``modeling a probabilistic
grammar''. So, we are convinced that formal-grammar writers should
help probabilistic-grammar modelers, and the other way round.

To exemplify this, we will have a closer look at the examples above,
where we presented two common types of ambiguity. In general, a PCFG
resolves ambiguity (i) by calculating all the full parse-trees of a
given sentence (using the symbolic backbone of the CFG), and (ii) by
allocating probabilities to all these trees (using the rule
probabilities of the PCFG), and finally (iii) by choosing the most
probable parse as the analysis of the given sentence.  According to
this procedure, we are calculating, for example
\begin{center}
\treesize{    
\Tree [.S \qroof{Peter}.NP [.VP [.V saw ] \qroof{Mary}.NP \qroof{with
a telescope}.PP ] ]
	
\Tree [.S \qroof{Peter}.NP [.VP [.V saw ] [.NP \qroof{Mary}.NP \qroof{with
a telescope}.PP ] ] ] 
}		
\end{center}
\begin{tabular}{cc}
\hspace{-1ex}\fbox{\begin{minipage}{7.5cm}
\begin{displaymath}
\begin{array}{l}
p(\treesize{S \rewr NP VP}) \cdot
p\left(\treesize{\qroof{Peter}.NP} \right) \cdot\\
p(\treesize{\textbf{VP \rewr V NP PP}}) \cdot \\
p(\treesize{V \rewr saw}) \cdot
p\left(\treesize{\qroof{Mary}.NP} \right) \cdot
p\left(\treesize{\qroof{with a telescope}.PP} \right) 
\end{array}
\end{displaymath}
\end{minipage}}
&
\hspace{-1ex}\fbox{\begin{minipage}{7.5cm}
\begin{displaymath}
\begin{array}{l}
p(\treesize{S \rewr NP VP}) \cdot
p\left(\treesize{\qroof{Peter}.NP} \right) \cdot\\
p(\treesize{\textbf{VP \rewr V NP}}) \cdot
p(\treesize{\textbf{NP \rewr NP PP}})\\
p(\treesize{V \rewr saw}) \cdot
p\left(\treesize{\qroof{Mary}.NP} \right) \cdot
p\left(\treesize{\qroof{with a telescope}.PP} \right) 
\end{array}
\end{displaymath}
\end{minipage}}
\end{tabular}
\ \\\\
\noindent
Comparing the probabilities for these two analyzes, we are
choosing the analysis at the left-hand side of this figure, if
\begin{displaymath}
	p(\treesize{\textbf{VP \rewr V NP PP}}) > 
	p(\treesize{\textbf{VP \rewr V NP}}) \cdot
p(\treesize{\textbf{NP \rewr NP PP}})
\end{displaymath}
So, in principle, the PP-attachment ambiguity encoded in this CFG
\textbf{can} be solved by a probabilistic model built on top of this
CFG. Moreover, it is especially nice that such a PCFG resolves the
ambiguity by looking only at those rules of the CFG, which directly
cause the PP-attachment ambiguity.

So far, so good: We are able to use PCFGs in order to select between
competing analyzes of a sentence. Looking at the second example
(ambiguity caused by conjunctions), however, we are faced with a
serious problem: Both trees have a different structure, but exactly
the same context-free rules are used for generating these different
structures. As a consequence, both trees are allocated the same
probability (independently from the specific rule probabilities which
might have been offered to you by the very best estimation
methods). So, any PCFG based on the given CFG is unable to resolve the
ambiguity manifested in the two trees.

Here is another problem. Using the grammar underlying our first
example, the sentence ``the girl saw a bird on a tree'' has the
following two analyzes
\begin{center}
\treesize{    	
\Tree [.S \qroof{the girl}.NP [.VP [.V saw ] [.NP \qroof{a bird}.NP
\qroof{on a tree}.PP ] ] ]

\Tree [.S \qroof{the girl}.NP [.VP [.V saw ] \qroof{a bird}.NP
\qroof{on a tree}.PP ] ] 
}		
\end{center}
Comparing the probabilities for these two analyzes, we are
choosing the analysis at the left-hand side of this figure, if
\begin{displaymath}
	p(\treesize{\textbf{VP \rewr V NP}}) \cdot
	p(\treesize{\textbf{NP \rewr NP PP}}) >
	p(\treesize{\textbf{VP \rewr V NP PP}})
\end{displaymath}
Relating this result to the disambiguation result of the sentence
``John saw Mary with a telescope'', the prepositional phrases are
attached in both cases either to the verbal phrase or to the nominal
phrase. Instead, it seems to be more plausible that the PP ``with the
telescope'' is attached to the verbal phrase, whereas the PP ``on a
tree'' is attached to the noun phrase.

Obviously, there is only one possible solution to this problem: We
have to re-write the given CFG in such a way that a probabilistic
model will be able to assign different probabilities to different
analyzes. For our last example, it is sufficient to enrich the
underlying CFG with some simple PP markup, enabling in principle that
\begin{eqnarray*}
	p(\treesize{\textbf{VP \rewr V NP PP-ON}}) 
&<&
	p(\treesize{\textbf{VP \rewr V NP}}) \cdot
	p(\treesize{\textbf{NP \rewr NP PP-ON}})
\\
	p(\treesize{\textbf{VP \rewr V NP PP-WITH}}) 
&>&
	p(\treesize{\textbf{VP \rewr V NP}}) \cdot
	p(\treesize{\textbf{NP \rewr NP PP-WITH}})
\end{eqnarray*}
Of course, other linguistically more sophisticated modifications of
the original CFG (that handle e.g. agreement information, sub-cat
frames, selectional preferences, etc) are also welcome. Our only
request is that the modified CFGs lead to PCFGs which are able to
resolve the different types of ambiguities encoded in the original
CFG.  Now, writing and re-writing a formal grammar is a job that
grammar writers can do probably much better than modelers of
probabilistic grammars.  In the past, however, writers of formal
grammars seemed to be uninterested in this specific task, or they are
still unaware of its existence. So, modelers of PCFGs regard it
nowadays also as an important part of their job to transform a given
CFG in such a way that probabilistic versions of the modified CFG are
able to resolve ambiguities of the original CFG.  During the last
years, a bunch of automatic grammar-transformation techniques have
been developed, which offer some interesting solutions to this quite
complex problem. Where the work of \newcite{KleinManning:2003}
describes one of the latest approaches to semi-automatic
grammar-transformation, the
\textbf{parent-encoding technique} introduced by
\newcite{Johnson:1998}
is the earliest and purely automatic grammar-transformation technique:
For each local tree, the parent's category is appended to all daughter
categories.
Using the example above, where we showed that
conjunctions cause ambiguities, the parent-encoded trees are looking
as follows:
\vspace{-3ex}
\begin{center}
\treesize{    
\hspace{-6ex}
\Tree [.S [.NP\append{S}  \qroof{the mother}.NP\append{NP} [.PP\append{NP} [.P\append{PP} of ] [.NP\append{PP} \qroof{the
boy}.NP\append{NP} [.CONJ\append{NP} and ] \qroof{the girl}.NP\append{NP} ] !\nodesize{24ex} ] ]
!\nodesize{60ex} \qroof{left}.VP\append{S} ]
\hspace{-12ex}
\Tree [.S [.NP\append{S} [.NP\append{NP}  \qroof{the mother}.NP\append{NP} [.PP\append{NP} [.P\append{PP} of ] \qroof{the
boy}.NP\append{PP} ] ]
!\nodesize{20ex} [.CONJ\append{NP} and ] \qroof{the
girl}.NP\append{NP} ]
!\nodesize{40ex} \qroof{left}.VP\append{S} ]
}
\end{center}
Clearly, parent-encoding of the original trees may result in different
probabilities of the transformed trees: In this example, we will
choose the analysis at the left-hand side, if
\begin{displaymath}
	p(\treesize{\textbf{NP\append{S}} \rewr NP\append{NP} PP\append{NP}})
	\cdot
	p(\treesize{\textbf{NP\append{PP}} \rewr NP\append{NP} CONJ\append{NP} NP\append{NP}})
	\cdot
	p\left(\treesize{\qroof{the boy}.\textbf{NP\append{NP}}}\right)
\end{displaymath} 
is more likely than
\begin{displaymath}
	p(\treesize{\textbf{NP\append{NP}} \rewr NP\append{NP} PP\append{NP}})
	\cdot
	p(\treesize{\textbf{NP\append{S}} \rewr NP\append{NP} CONJ\append{NP}
	NP.\append{NP}})
	\cdot
	p\left(\treesize{\qroof{the boy}.\textbf{NP\append{PP}}}\right)
\end{displaymath}
As in the example before, it is again nice to see that these
probabilities are pin-pointing exactly at those rules of the
underlying grammar which have introduced the ambiguity.

In the rest of the section, we will present the notion of treebank
grammars, which can be informally described as PCFGs that are
constructed on the basis of a corpus of full-parse trees
\cite{Charniak:96}.  We will demonstrate that treebank
grammars can resolve the ambiguous sentences of the treebank (as well
as ambiguous similar sentences), if the treebank mark-up is rich
enough to distinguish between the different types of ambiguities that
are encoded in the treebank.

\begin{definition}
\label{treebank.training}
For a given \textbf{treebank}, i.e., for a non-empty and finite corpus
of full-parse trees, the \textbf{treebank grammar} $<G,p>$ is a PCFG
defined by
\begin{itemize}
\item[(i)] $G$ is the context-free grammar read off from the
treebank, and
\item[(ii)] $p$ is the probability distribution on the set
of full-parse trees of $G$, induced by the following specific
probability distributions on the grammar fragments $G_A$:
\begin{displaymath}
	p(r) = \frac{f(r)}{\sum_{r \in G_A} f(r)}
	\qquad
	\mbox{ for all } r \in G_A
\end{displaymath} 
Here, $f(r)$ is the number of times a rule $r \in G$ occurs in the
treebank.
\end{itemize}
\end{definition}
\textbf{Note:} Later on (see Theorem~\ref{theorem.treebank.training}), 
we will show that each treebank grammar is the unique
maximum-likelihood estimate of $G$'s probability model on the given
treebank. So, it is especially guaranteed that $p$ is a probability
distribution on the set of full-parse trees of $G$, and that $<G,P>$
is a standard PCFG (see Definition~\ref{def.pcfg}).

\begin{example}
\label{example.treebank.grammar}
We shall now consider a treebank given by the following 210 full-parse
trees:
\ \\\\
\treesize{    
\textbf{\underline{100 $\times\ t_1$:}}\hspace{-10ex}
\Tree [.S \qroof{Peter}.NP [.VP [.V saw ] \qroof{Mary}.NP \qroof{with
a telescope}.PP-WITH ] ]

\textbf{\underline{5 $\times\ t_2$:}}\hspace{-10ex}	
\Tree [.S \qroof{Peter}.NP [.VP [.V saw ] [.NP \qroof{Mary}.NP \qroof{with
a telescope}.PP-WITH ] ] ]
\ \\\\
\textbf{\underline{100 $\times\ t_3$:}}\hspace{-7ex}	
\Tree [.S \qroof{Mary}.NP [.VP [.V saw ] [.NP \qroof{a bird}.NP
\qroof{on a tree}.PP-ON ] ] ]

\textbf{\underline{5 $\times\ t_4$:}}\hspace{-7ex}
\Tree [.S \qroof{Mary}.NP [.VP [.V saw ] \qroof{a bird}.NP \qroof{on a
tree}.PP-ON ] ] 
}

\noindent
We shall (i) generate the treebank grammar and (ii) using this
treebank grammar, we shall resolve the ambiguities of the sentences
occurring in the treebank.
\end{example}
Ad (i): The following table displays the rules $r$ of the CFG encoded
in the given treebank (thereby assuming for the sake of simplicity
that the NP and PP non-terminals expand directly to terminal symbols),
the rule frequencies $f(r)$ i.e. the number of times a rule occurs in
the treebank, as well as the rule probabilities $p(r)$ as defined in
Definition~\ref{treebank.training}.

\begin{center}
\begin{tabular}{l|l|l}
\textbf{CFG rule} & \textbf{Rule frequency} &
\textbf{Rule probability}
\\
\hline
\treesize{S \rewr NP VP} 	& 100 + 5 + 100 + 5 & 
				\treesize{$\frac{210}{210} = 1.000$}\\
&&\\
\treesize{VP \rewr V NP PP-WITH} 	& 100 & 
				\treesize{$\frac{100}{210} \approx 0.476$}\\
\treesize{VP \rewr V NP PP-ON} & 5 & 
				\treesize{$\frac{5}{210} \approx 0.024$}\\
\treesize{VP \rewr V NP} & 5 + 100 & 
				\treesize{$\frac{105}{210} = 0.500$}\\
&&\\
\treesize{NP \rewr Peter} & 100 + 5 & \treesize{$\frac{105}{525} = 0.200$}\\
\treesize{NP \rewr Mary} & 100 + 5 + 100 + 5 & \treesize{$\frac{210}{525} = 0.400$}\\
\treesize{NP \rewr a bird} & 100 + 5 & \treesize{$\frac{105}{525} = 0.200$}\\
\treesize{NP \rewr NP PP-WITH} & 5  & \treesize{$\frac{5}{525} \approx 0.010$}\\
\treesize{NP \rewr NP PP-ON} & 100  & \treesize{$\frac{100}{525} \approx 0.190$}\\
&&\\
\treesize{PP-WITH \rewr with a telescope} & 100 + 5 & \treesize{$\frac{105}{105} = 1.000$}\\
&&\\
\treesize{PP-ON \rewr on a tree} & 100 + 5 & \treesize{$\frac{105}{105} = 1.000$}\\
&&\\
\treesize{V \rewr saw} & 100 + 5 + 100 + 5 & \treesize{$\frac{210}{210} = 1.000$}\\
\end{tabular}
\end{center}
Ad (ii): As we have already seen, the treebank grammar selects the
full-parse tree $t_1$ of the sentence ``Peter saw Mary with a
telescope'', if
\begin{displaymath}
	p(\treesize{\textbf{VP \rewr V NP PP-WITH}}) 
	>
	p(\treesize{\textbf{VP \rewr V NP}}) \cdot
	p(\treesize{\textbf{NP \rewr NP PP-WITH}})
\end{displaymath}
Using the approximate probabilities for these rules, this is indeed
true: $0.476 > 0.500 \cdot 0.010$. (Note that exactly the same
argument can be applied to similar but more complex sentences like
``Peter saw a bird on a tree with a telescope''.)  For the second
sentence occurring in the treebank, ``Mary saw a bird on a tree'', the
treebank grammar selects the full-parse tree $t_3$, if
\begin{displaymath}
	p(\treesize{\textbf{VP \rewr V NP PP-ON}}) 
	<
	p(\treesize{\textbf{VP \rewr V NP}}) \cdot
	p(\treesize{\textbf{NP \rewr NP PP-ON}})
\end{displaymath}
Indeed, this is the case: $0.024 < 0.500 \cdot 0.190$.

\section*{Maximum-Likelihood Estimation of PCFGs}

\begin{figure*}
\begin{center}
\mbox{\includegraphics[width=0.90\textwidth]{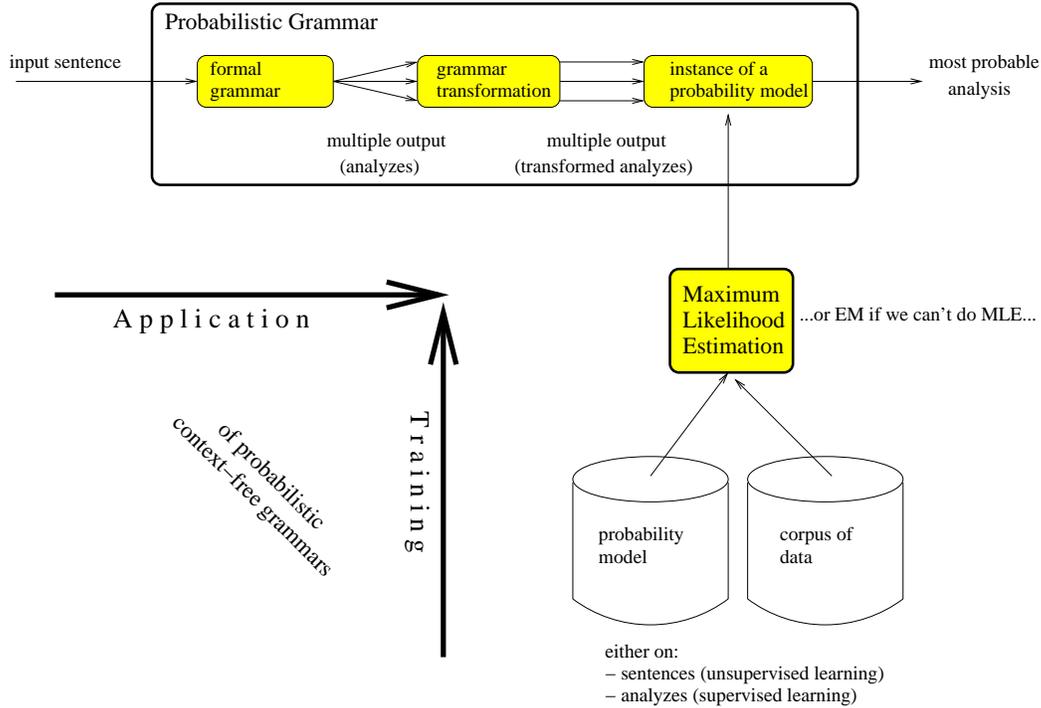}}
\end{center}
\caption{{\small Application and training of probabilistic
  context-free grammars}}
\label{figure.overview}
\end{figure*}

So far, we have seen that probabilistic context-free grammars can be
used to resolve the ambiguities that are caused by their underlying
context-free backbone, and we noted already that certain
grammar-transformation are sometimes necessary to achieve this
goal. All these application features are displayed in a ``horizontal
view'' in Figure~\ref{figure.overview}. In what follows next, we will
concentrate on the ``vertical view'' of this figure. To be more
specific, we will focus on the following two questions.
\begin{itemize}
\item[(i)]how to characterize
the probability model of a given context-free grammar, and 
\item[(ii)]second, how to estimate an appropriate instance of the
context-free grammar's probability model, if a corpus of input data is
additionally given.
\end{itemize}
The latter question is a tough one: It is true that the
treebank-training method, which we defined in the previous section
more or less heuristically, leads to PCFGs that are able to resolve
ambiguities. From what we have done so far in this section, however,
we have no clear idea how the treebank-training method is related to
maximum-likelihood estimation or the EM training method. So, let us
start with the first question.

\begin{definition}
\label{def.cfg.model}
Let $G$ be a context-free grammar, and let $\mathcal{X}$ be the set of
full-parse trees of $G$. Then, the \textbf{probability model of
$G$} is defined by
\begin{displaymath}
  \mathcal{M}_G = \left\{ p \in \mathcal{M}(\mathcal{X}) \left| \ p(x)
  = \prod_{r \in G} p(r)^{f_r(x)} \mbox{ with } \sum_{r \in G_A} p(r)
  = 1 \mbox{ for all grammar fragments } G_A \right\}\right.
\end{displaymath}
In other words, each instance $p$ of the probability model
$\mathcal{M}_G$ is associated to a probabilistic context-free grammar
$<G,p>$ having the context-free backbone $G$. (See
Definition~\ref{def.pcfg} for the meaning of the terms $p(r)$,
$f_r(x)$ and $G_A$.)
\end{definition}
As we have already seen, there are some non-standard PCFGs (like
\mbox{\treesize{S \rewr S S (0.9), S \rewr a (0.1)}\hspace{-1ex})}
which do not induce a probability distribution on the set of
full-parse trees. This gives us a rough idea that it might be quite
difficult to characterize those PCFGs $<G,p>$ which are associated to
an instance $p$ of the unrestricted probability model
$\mathcal{M}(\mathcal{X})$. In other words, it might be quite
difficult to characterize $G$'s probability model $\mathcal{M}_G$. For
calculating a maximum-likelihood estimate of $\mathcal{M}_G$ on a
corpus $f_{\mathcal{T}}$ of full-parse trees, however, we have to
solve this task. For example, if we are targeting to exploit the
powerful Theorem~\ref{th.mle}, we have to prove either that
$\mathcal{M}_G$ equals the unrestricted probability model
$\mathcal{M}(\mathcal{X})$, or that the relative-frequency estimate
$\tilde{p}$ on $f_{\mathcal{T}}$ is an instance of
$\mathcal{M}_G$. For most context-free grammars $G$, however, the
following theorems show that this is a too simplistic approach of
finding a maximum-likelihood estimate of $\mathcal{M}_G$.

\begin{theorem}
\label{theorem.pcfg.relfreq}
Let G be a context-free grammar, and let $\tilde{p}$ be the
relative-frequency estimate on a non-empty and finite corpus
$f_{\mathcal{T}}\!: \mathcal{X} \rightarrow \mathcal{R}$ of full-parse
trees of $G$. 
Then $\tilde{p} \not\in
\mathcal{M}_G$ if
\begin{itemize}
\item[(i)] $G$ can be read off from $f_{\mathcal{T}}$, and
\item[(ii)] $G$ has a full-parse tree $x_\infty \in \mathcal{X}$ that
is not in $f_{\mathcal{T}}$.
\end{itemize}
\end{theorem}

\proof Assume that $\tilde{p} \in \mathcal{M}_G$. In what follows, we
will show that this assumption leads to a contradiction. First, by
definition of $\mathcal{M}_G$, it follows that there are some weights
$0 \le \pi(r) \le 1$ such that
\begin{displaymath}
	\tilde{p}(x) = \prod_{r \in G} \pi(r)^{f_r(x)} 
	\qquad\mbox{ for all } x \in \mathcal{X}
\end{displaymath}
We will show next that $\pi(r) > 0$ for all $r \in G$.
\begin{quote}
Assume that there is a rule $r_0 \in G$ with $\pi(r_0) = 0$. By (i),
$G$ can be read off from $f_{\mathcal{T}}$. So, $f_{\mathcal{T}}$
contains a full-parse tree $x_0$ such that the rule $r_0$
occurs in $x_0$, i.e.,
\begin{displaymath}
	f_{\mathcal{T}}(x_0) > 0 
	\quad\mbox{ and }\quad
	f_{r_0}(x_0) > 0
\end{displaymath}
It follows both $\tilde{p}(x_0) =
\frac{f_{\mathcal{T}}(x_0)}{|f_{\mathcal{T}}|} > 0$ and 
$\tilde{p}(x_0) = \prod_{r \in G} \pi(r)^{f_r(x_0)} = 
\cdots \pi(r_0)^{f_{r_0}(x_0)} \cdots = 0$, which is a
 contradiction.
\end{quote}
Therefore
\begin{displaymath}
	\tilde{p}(x) = \prod_{r \in G} \pi(r)^{f_r(x)} > 0 
	\qquad\mbox{ for all } x \in \mathcal{X}
\end{displaymath}
On the other hand by (ii), there is the full-parse tree $x_\infty \in
\mathcal{X}$ which is not in $f_{\mathcal{T}}$. So,
$\tilde{p}(x_\infty)=\frac{f_{\mathcal{T}}(x_\infty)}{|f_{\mathcal{T}}|}
= 0$, which is a contradiction to the last inequality \textbf{q.e.d.}

\begin{example}
The treebank grammar described in
Example~\ref{example.treebank.grammar} is a context-free grammar of
the type described in Theorem~\ref{theorem.pcfg.relfreq}.
\end{example}
The relative-frequency estimate $\tilde{p}$ on the treebank is given
by:
\begin{displaymath}
  \tilde{p}(t_1) = \tilde{p}(t_3) = \frac{100}{210}
  \qquad\mbox{ and }\qquad
  \tilde{p}(t_2) = \tilde{p}(t_4) = \frac{5}{210}
\end{displaymath}
All other full-parse trees of the treebank grammar get allocated a
probability of zero by the relative-frequency estimate. So, for
example, $\tilde{p}(x_\infty)=0$ for
\begin{center}
\treesize{    
\textbf{\underline{$x_\infty$:}}\hspace{-10ex}
\Tree [.S \qroof{Mary}.NP [.VP [.V saw ] \qroof{Peter}.NP \qroof{with
a telescope}.PP-WITH ] ]}
\end{center}
As a consequence, $\tilde{p}$ can not be an instance of the
probability model of the treebank grammar: Otherwise, $\tilde{p}$
would allocate both full-parse trees $t_1$ and $x_\infty$ exactly the
same probabilities (because $t_1$ and $x_\infty$ contain exactly the
same rules).

\begin{theorem}
For each context-free grammar $G$ with an infinite set $\mathcal{X}$
of full-parse trees, the probability model of $G$ is restricted
\begin{displaymath}
	\mathcal{M}_G \not= \mathcal{M}(\mathcal{X})
\end{displaymath}
\end{theorem}

\proof First, each context-free grammar consists of a finite number of
rules. Thus it is possible to construct a treebank, such that $G$ is
encoded by the treebank. (Without loss of generality, we are assuming
here that all non-terminal symbols of $G$ are reachable and productive.) So,
let $f_{\mathcal{T}}$ be a non-empty and finite corpus of full-parse
trees of $G$ such that $G$ can be read off from $f_{\mathcal{T}}$, and
let $\tilde{p}$ be the relative-frequency estimate on
$f_{\mathcal{T}}$. Second, using $|\mathcal{X}| =
\infty$ and $|f_{\mathcal{T}}| < \infty$, it follows that there is at
least one full-parse tree $x_\infty \in
\mathcal{X}$ which is not in $f_{\mathcal{T}}$. So, the conditions
of Theorem~\ref{theorem.pcfg.relfreq} are met, and we are concluding
that $\tilde{p} \not\in \mathcal{M}_G$. On the other hand, $\tilde{p}
\in \mathcal{M}(\mathcal{X})$. So, clearly, $\mathcal{M}_G \not=
\mathcal{M}(\mathcal{X})$. In other words, $\mathcal{M}_G$ is a
restricted probability model \textbf{q.e.d.}
\\
\\
\noindent
After all, we might recognize that the previous results are not bad.
Yes, the probability model of a given CFG is restricted in most of the
cases. The missing distributions, however, are the relative-frequency
estimates on each treebank encoding the given CFG. These
relative-frequency estimates lack the ability of any generalization
power: They allocate each full-parse tree not being in the treebank a
zero-probability.  Obviously, however, we surely want a probability
model that can be learned by maximum-likelihood estimation on a corpus
of full-parse trees, but that is at the same time able to deal with
full-parse trees not seen in the treebank. The following theorem shows
that we have already found one.

\begin{theorem}
\label{theorem.treebank.training}
Let $f_{\mathcal{T}}\!: \mathcal{X} \rightarrow \mathcal{R}$ be a
non-empty and finite corpus of full-parse trees, and let
$<G,p_{\mathcal{T}}>$ be the treebank grammar read off from
$f_{\mathcal{T}}$. Then, $p_{\mathcal{T}}$ is a maximum-likelihood
estimate of $\mathcal{M}_G$ on $f_{\mathcal{T}}$, i.e.,
\begin{displaymath}
   L(f_{\mathcal{T}}; p_{\mathcal{T}}) = 
   \max_{p \in \mathcal{M}_G} L(f_{\mathcal{T}}; p)
\end{displaymath}
Moreover, maximum-likelihood estimation of $\mathcal{M}_G$ on
$f_{\mathcal{T}}$ yields a unique estimate.
\end{theorem}
This theorem is well-known. The following proof, however, is
especially simple and (to the best of my knowledge) was given first by
\newcite{Prescher:Diss}.
\\
\\
\proof \underline{First step:} We will show that for all model instances
$p \in \mathcal{M}_G$
\begin{displaymath}
  L(f_{\mathcal{T}}; p) = L(f; p) 
\end{displaymath}
At the right-hand side of this equation, $f$ refers to the corpus of
rules that are read off from the treebank $f_\mathcal{T}$, i.e.,
$f(r)$ is the number of times a rule $r \in G$ occurs in the treebank;
Similarly, $p$ refers to the probabilities $p(r)$ of the rules $r
\in G$. In contrast, at the left-hand side of
the equation, $p$ refers to the probabilities $p(x)$ of the full-parse
trees $x \in \mathcal{X}$. The proof of the equation is relatively
straight-forward:

\begin{eqnarray*} 
  L(f_{\mathcal{T}}; p)
  &=&
  \prod_{x \in \mathcal{X}} p(x)^{f_{\mathcal{T}}(x)}
\\
  &=&
  \prod_{x \in \mathcal{X}} 
  \left(
  \prod_{r \in G} p(r)^{f_r(x)}
  \right)^{f_{\mathcal{T}}(x)}
\\
  &=&
  \prod_{x \in \mathcal{X}} \prod_{r \in G} 
  p(r)^{f_{\mathcal{T}}(x) \cdot f_r(x)}
\\
  &=&
  \prod_{r \in G}  \prod_{x \in \mathcal{X}} 
  p(r)^{f_{\mathcal{T}}(x) \cdot f_r(x)}
\\
  &=&
  \prod_{r \in G}
  p(r)^{\sum_{x \in \mathcal{X}} f_{\mathcal{T}}(x) \cdot f_r(x)}
\\
  &=&
  \prod_{r \in G}
  p(r)^{f(r)}
\\
  &=&
  L(f; p)
\end{eqnarray*}
In the $6^{th}$ equation, we simply used that $f(r)$ (the number of
times a specific rule occurs in the treebank) can be calculated by
summing up all the $f_r(x)$ (the number of times this rule occurs in a
specific full-parse tree~$x \in \mathcal{X}$):
\begin{displaymath}
  f(r) = \sum_{x \in \mathcal{X}} f_{\mathcal{T}}(x) \cdot f_r(x)
\end{displaymath}
So, maximizing $L(f_{\mathcal{T}}; p)$ is equivalent to maximizing
$L(f; p)$. Unfortunately, Theorem~\ref{th.mle} can not be applied to
maximize the term $L(f;p)$, because the rule probabilities $p(r)$ do not form
a probability distribution on the set of all grammar rules. They do
form, however, probability distributions on the grammar fragments
$G_A$. So, we have to refine our result a bit more.  \ \\\\
\noindent 
\underline{Second step:} We are showing here that
\begin{displaymath}
   L(f; p) = \prod_A L(f_A; p)
\end{displaymath}
Here, each $f_A$ is a corpus of rules, read off from the given
treebank, thereby filtering out all rules not having the left-hand
side $A$. To be specific, we define
\begin{displaymath}
  f_A(r)
  =
  \left\{
  \begin{array}{cll}
    f(r) & \mbox{ if }   & r \in G_A\\
    0    & \mbox{ else } &\\
  \end{array}
  \right.
\end{displaymath}
Again, the proof is easy:
\begin{eqnarray*}
  L(f; p) 
  &=& 
  \prod_{r \in G} p(r)^{f(r)}
\\  
  &=& 
  \prod_A \prod_{r \in G_A} p(r)^{f(r)}
\\
  &=& 
  \prod_A \prod_{r \in G_A} p(r)^{f_A(r)}
\\
  &=& 
  \prod_A L(f_A; p)
\end{eqnarray*}
\underline{Third step:} Combining the first and second step, we
conclude that
\begin{displaymath}
  L(f_{\mathcal{T}}; p) = \prod_{A} L(f_A; p) 
\end{displaymath}
So, maximizing $L(f_{\mathcal{T}}; p)$ is equivalent to maximizing
$\prod_{A} L(f_A; p)$. Now, a product is maximized, if all its factors
are maximized. So, in what follows, we are focusing on how to
maximize the terms $L(f_A; p)$. First of all, the corpus $f_A$
comprises only rules with the left-hand side A. So, the value of
$L(f_A;p)$ depends only on the values $p(r)$ of the rules $r \in
G_A$. These values, however, form a probability distribution on $G_A$,
and all these probability distributions on $G_A$ have to be considered
for maximizing $L(f_A;p)$. It follows that we have to calculate an
instance $\hat{p}_A$ of the unrestricted probability model
$\mathcal{M}(G_A)$ such that
\begin{displaymath}
  L(f_A;\hat{p}_A) = \max_{p \in \mathcal{M}(G_A)} L(f_A;p)
\end{displaymath}
In other words, we have to calculate a maximum-likelihood estimate of
the unrestricted probability model $\mathcal{M}(G_A)$ on the corpus
$f_A$. Fortunately, this task can be easily solved. According to
Theorem~\ref{th.mle}, the relative-frequency estimate on the corpus
$f_A$ is our unique solution. This yields for the rules $r \in G_A$
\begin{displaymath}
  \hat{p}_A(r) 
  = \frac{f_A(r)}{|f_A|}
  = \frac{f(r)}{\sum_{r \in G_A} f(r)}
\end{displaymath}
Comparing these formulas to the formulas given in
Definition~\ref{treebank.training}, we conclude that for all
non-terminal symbols $A$
\begin{displaymath}
  \hat{p}_A(r) 
  = 
  p_{\mathcal{T}}(r)
  \quad
  \mbox{ for all }
  r \in G_A
\end{displaymath}
So, clearly, the treebank grammar $p_{\mathcal{T}}$ is a serious
candidate for a maximum-likelihood estimate of the probability model
$\mathcal{M}_G$ on $f_{\mathcal{T}}$. Now, as \newcite{Chi:1998b}
verified, the treebank grammar $p_{\mathcal{T}}$ is indeed an instance
of the probability model $\mathcal{M}_G$.  So, combining all the
results, it follows that
\begin{displaymath}
  L(f_{\mathcal{T}}; p_{\mathcal{T}}) = \max_{p \in \mathcal{M}_G}
  L(f_{\mathcal{T}}; p)
\end{displaymath}
Finally, since all $\hat{p}_A \in \mathcal{M}(G_A)$ are unique
maximum-likelihood estimates, $p_{\mathcal{T}} \in \mathcal{M}_G$ is
also a unique maximum-likelihood estimate \textbf{q.e.d.}

\section*{EM Training of PCFGs}

Let us present first an overview of some theoretical work on using the
EM algorithm for training of probabilistic context-free grammars.

\begin{itemize}
\item From 1966 to 1972, a group around Leonard E. Baum invents the
  forward-backward algorithm for probabilistic regular grammars
  (or hidden Markov models) and formally proves that this algorithm has
  some good convergence properties. See, for example, the overview
  presented in \newcite{Baum:72}.
\item \newcite{Booth:73} discover a (still nameless) constraint for
  PCFGs. For PCFGs fulfilling the constraint, the probabilities
  of all full-parse trees sum to one.
\item[$\circ$] \textbf{\newcite{Dempster:77} invent the EM
  algorithm.}
\item \newcite{Baker:79} invents the inside-outside algorithm (as a
  generalization of the forward-backward algorithm). The training
  corpus of this algorithm, however, is not allowed to contain more
  than one single sentence.
\item \newcite{Lari:90} generalize Baker's inside-outside algorithm
  (or in other words, they invent the modern form of the
  inside-outside algorithm): The training corpus of their algorithm
  may contain arbitrary many sentences.
\item \newcite{Pereira:92} use the inside-outside algorithm for
  estimating a PCFG for English on a partially bracketed corpus.
\item \newcite{Sanchez1997} and \newcite{Chi:1998b} formally prove
  that for treebank grammars and grammars estimated by the EM
  algorithm, the probabilities of all full-parse trees sum to one.
\item \newcite{Prescher:01c} formally proves that the inside-outside
  algorithm is a dynamic programming instance of the EM algorithm for
  PCFGs. As a consequence, the inside-outside algorithm inherits
  the convergence properties of the EM algorithm (no formal
  proofs of these properties have been given by Baker and Lari and
  Young).
\item \newcite{NederhofSatta:2003} discover that the PCFG
  standard-constraints (``the probabilities of the rules with the same
  left-hand side are summing to one'') are dispensable
  \footnote{In our terms, their result can be expressed as follows.
    Let $G$ be a context-free grammar, and let $\mathcal{M}_G^*$ be
    the probability model that disregards the PCFG
    standard-constraints
    \[
      \mathcal{M}_G^* = \left\{ p \in \mathcal{M}(\mathcal{X}) \left|
      \ p(x) = \prod_{r \in G} p(r)^{f_r(x)} \right\}\right.
    \]
    Obviously, it follows then that $\mathcal{M}_G \subseteq
    \mathcal{M}_G^*$.  Exploiting their Corollary~8, however, it
    follows somewhat surprisingly that both models are even equal:
    $\mathcal{M}_G^* = \mathcal{M}_G$. As a trivial consequence, a
    maximum-likelihood estimate of the standard probability model
    $\mathcal{M}_G$ (on a corpus of trees or on a corpus of sentences)
    is also a maximum-likelihood estimate of the probability model
    $\mathcal{M}_G^*$ --- as well as the other way round.}.
\end{itemize}
The overview presents two interesting streams. First, it suggests that
the forward-backward algorithm is a special instance of the
inside-outside algorithm, which in turn is a special instance of the
EM algorithm. Second, it appears to be worthy to reflect on our notion
of a standard probability model for context-free grammars (see the
results of \newcite{Booth:73} and
\newcite{NederhofSatta:2003}). Surely, both topics are very
interesting --- Here, however, we would like to limit our selfs and
refer the interested reader to the papers mentioned above. The rest of
this paper is dedicated to the pure non-dynamic EM algorithm for
PCFGs. We would like to present its procedure and its properties,
thereby motivating that the EM algorithm can be successfully used to
train a manually written grammar on a plain text corpus.

As a first step, the following theorem shows that the EM algorithm is
not only related to the inside-outside algorithm, but that the EM
algorithm is also strongly connected with the treebank-training method
on which we we have focused so far.

\begin{theorem}
\label{theorem.em.training}
Let $<G, p_0>$ be a probabilistic context-free grammar, where $p_0$ is
an arbitrary starting instance of the probability model
$\mathcal{M}_G$. Let $f\!: \mathcal{Y} \rightarrow \mathcal{R}$ be a
non-empty and finite corpus of \textbf{sentences of $G$} (terminal
strings that have at least one full-parse tree $x \in
\mathcal{X}$). Then, applied to the PCFG $<G, p_0>$ and the sentence
corpus $f$, the EM Algorithm performs the following procedure\ \\\\
\noindent 
(1)\hspace{1.0cm} for each $i=1,\ 2,\ 3,\ ...$ do 
\ \\
(2)\hspace{1.5cm} $q := p_{i-1}$ 
\ \\ 
(3)\hspace{1.5cm} \textbf{E-step (PCFGs):} generate the \textbf{treebank}
   $f_{\mathcal{T}_q}\!: \mathcal{X} \rightarrow \mathcal{R}$ defined by
\begin{eqnarray*}
  &&\qquad f_{\mathcal{T}_q}(x) \ := f(y) \cdot q(x|y) \quad \mbox{ where
  } y = \mbox{yield}(x)
\end{eqnarray*}
(4)\hspace{1.5cm} \textbf{M-step (PCFGs):} read off the
  \textbf{treebank grammar} $<G, p_{\mathcal{T}_q}>$\ \\
(5)\hspace{1.5cm} $p_i := p_{\mathcal{T}_q}$
\ \\
(6)\hspace{1.0cm} end // for each $i$ 
\ \\
(7)\hspace{1.0cm} print $p_0, p_1, p_2, p_3, ...$ \
\ \\\\
\noindent
Moreover, these EM re-estimates allocate the corpus $f$ a sequence of
corpus probabilities that is monotonic increasing
\begin{displaymath}
  L(f; p_0) \le L(f; p_1) \le L(f; p_2) \le L(f; p_3) \le \ldots
\end{displaymath}
\end{theorem}

\proof See Definition~\ref{em.procedure}, and theorems~\ref{em.output}
and \ref{theorem.treebank.training}.
\\
\\
\noindent
Here is a toy example that exemplifies \textbf{how} the EM algorithm
is applied to PCFGs. Remind first that we showed in
Example~\ref{example.treebank.grammar} that a treebank grammar is (in
principle) able to disambiguate correctly different forms of PP
attachment. Remind also that we had to introduce some simple PP
mark-up to achieve this result. Although we are choosing here a
simpler example (so that the number of EM calculations is kept small),
the example shall provide us with an intuition about \textbf{why} the
EM algorithm is (in principle) able to learn ``good'' PCFGs. Again,
the toy example presents a CFG having a PP-attachment ambiguity. This
time, however, the training corpus is not made up of full-parse trees
but of sentences of the grammar. Two types shall occur in the given
corpus: Whereas the sentences of the first type are ambiguous, the
sentences of the second type are not. Our simple goal is to
demonstrate that the EM algorithm is able to learn from the
unambiguous sentences how to disambiguate the ambiguous ones.  It is
exactly this feature that enables the EM algorithm to learn highly
ambiguous grammars from real-world corpora: Although it is almost
guaranteed in practice that sentences have on average thousands of
analyzes, no sentence will hardly be completely unambiguous. Almost
all sentences in a given corpus contain partly unambiguous information
--- represented for example in small sub-trees that the analyzes of
the sentence share. By weighting the unambiguous information ``high'',
and at the same time, by weighting the ambiguous information ``low''
(indifferently, uniformly or randomly), the EM algorithm might output
a PCFG that learned something, namely, a PCFG being able to select for
almost all sentences the single analysis that fits best the
information which is hidden but nevertheless encoded in the corpus.

\begin{example}
We shall consider an experiment in which a manually written CFG is
estimated by the EM algorithm. We shall assume that the following
corpus of 15 sentences is given to us
\begin{center}
\begin{tabular}{cc}
\begin{tabular}{c|c}
  $f(y)$ & $y$\\
  \hline
  5 & $y_1$\\
  10 & $y2$\\
\end{tabular}
&
\begin{tabular}{l}
  $y_1$ = ``Mary saw a bird on a tree''\\
  $y_2$ = ``a bird on a tree saw a worm''\\
\end{tabular}
\end{tabular}
\end{center}
Using this text corpus, we shall compute the EM re-estimates of the
following PCFG (that is able to parse all the corpus sentences)
\begin{center}
\begin{tabular}{ll}
S \rewr NP VP & (1.00) \\
VP \rewr V NP & (0.50) \\
VP \rewr V NP PP & (0.50) \\  
NP \rewr NP PP & (0.25) \\
NP \rewr Mary & (0.25) \\ 
NP \rewr a bird & (0.25) \\
NP \rewr a worm & (0.25) \\
PP \rewr on a tree & (1.00) \\  
V \rewr saw & (1.00) \\
\end{tabular}
\end{center}
\textbf{Note:} As being displayed, the uniform distributions on the
grammar fragments ($G_S$, $G_{VP}$, $G_{NP}$ ...) serve in this
example as a starting instance $p_0 \in \mathcal{M}_G$. On the one
hand, this is not bad, since this is (or was) the common practice for
EM training of probabilistic grammars. One the other hand, the EM
algorithm gives us the freedom to experiment with a huge range of
starting instances. Now, why not using the freedom that the EM
algorithm donates? Indeed, the convergence proposition of
Theorem~\ref{theorem.em.training} permits that some starting instances
of the EM algorithm may lead to better re-estimates than other
starting instances. So, clearly, if we are experimenting with more
than one single starting instance, then we can seriously hope that our
efforts are re-compensated by getting a better probabilistic grammar!
\end{example}
First of all, note that the first sentence is ambiguous, whereas the
second is not. The following figure displays the full-parse trees of
both.  

\ \\ 
\treesize{ \textbf{\underline{$x_1$:}}\hspace{-7ex}
\Tree [.S \qroof{Mary}.NP [.VP [.V saw ] [.NP \qroof{a bird}.NP
\qroof{on a tree}.PP ] ] ]
\hspace{-12ex}
\textbf{\underline{$x_2$:}}\hspace{-7ex}
\Tree [.S \qroof{Mary}.NP [.VP [.V saw ] \qroof{a bird}.NP \qroof{on a
      tree}.PP ] ]
\hspace{-12ex}
\textbf{\underline{$x_3$:}}\hspace{-7ex}
\Tree [.S  [.NP \qroof{a bird}.NP
      \qroof{on a tree}.PP ] [.VP [.V saw ] \qroof{a worm}.NP ] ]
\vspace{-7ex}
}

\ \\\\
\noindent
\textbf{First EM iteration}. In the \textbf{E-step} of the EM
algorithm for PCFGs, a treebank $f_{\mathcal{T}_q}$ has to be
generated on the basis of the starting instance $q := p_0$.  The
generation process is very easy. First, we have to calculate the
probabilities of the full-parse trees $x_1, x_2$ and $x_3$, 
which in turn provide us with the
probabilities of the sentences $y_1$ and $y_2$.  Here are the results
\begin{displaymath}
\begin{array}{c|c}
  p_0(x) & x \\
\hline
0.0078125 &  x_1 \\
0.0312500 &  x_2 \\
0.0078125 &  x_3 \\
\end{array}
\mbox{\hspace{2cm}}
\begin{array}{c|c}
  p_0(y) & y \\
\hline
0.0390625 &  y_1 \\
0.0078125 &  y_2 \\
\end{array}
\end{displaymath}
For example, $p_0(x_1)$ and $p_0(y_1)$ are calculated as follows (using
definitions~\ref{def.pcfg} and~\ref{em.input}.*):
\begin{eqnarray*}
  p_0(x_1) 
  &=& 
  \prod_{r \in G} p(r)^{f_r(x_1)}
  \\
  &=&
  p(\treesize{S \rewr NP VP}) \cdot
  p\left(\mbox{\hspace{-1ex}}\treesize{\qroof{Mary}.NP}\mbox{\hspace{-1ex}}\right) \cdot
  p(\treesize{VP \rewr V NP}) \cdot
  p(\treesize{V \rewr saw}) \cdot
  p\left(\mbox{\hspace{-1ex}}
  \treesize{\qroof{a bird}.NP}\mbox{\hspace{-1ex}}\right) \cdot
  p\left(\mbox{\hspace{-1ex}}\treesize{\qroof{on a tree}.PP}\mbox{\hspace{-1ex}}\right) 
  \\
  &=&
  1.00 \cdot 0.25 \cdot 0.50 \cdot 1.00 \cdot 0.25 \cdot 0.25 \cdot
  1.00
  \\
  &=& 
  0.0078125
  \\\\
  p_0(y_1) 
  &=& 
  \sum_{yield(x)=y_1} p(x)
  \\
  &=&
  p(x_1) + p(x_2)
  \\
  &=&
  0.0078125 + 0.0312500
  \\
  &=&
  0.0390625
\end{eqnarray*}
Now, using the probability distribution $q := p_0$, we have to
generate the treebank $f_{\mathcal{T}_q}$ by distributing the
frequencies of the sentences to the full-parse trees. The result is
\begin{displaymath}
\begin{array}{c|c}
  f_{\mathcal{T}_q}(x) & x \\
\hline
1   &  x_1 \\
4  &  x_2 \\
10  &  x_3 \\
\end{array}
\end{displaymath}
For example, $f_{\mathcal{T}_q}(x_1)$ is calculated as follows (see
line (3) of the EM procedure in Theorem~\ref{theorem.em.training}):
\begin{eqnarray*}
  f_{\mathcal{T}_q}(x_1) 
  &=& 
  f(\mbox{yield}(x_1)) \cdot q(x_1|\mbox{yield}(x_1))
  \\ 
  &=&
  f(y_1) \cdot q(x_1|y_1)
  \\ 
  &=&
  f(y_1) \cdot \frac{q(x_1)}{q(y_1)}
  \\ 
  &=&
  5 \cdot \frac{0.0078125}{0.0390625}
  \\
  &=&
  1
\end{eqnarray*}
In the \textbf{E-step} of the EM algorithm for PCFGs, we have to read
off the treebank grammar $<G,p_{\mathcal{T}_q}>$ from the treebank
$f_{\mathcal{T}_q}$. Here is the result
\begin{center}
\begin{tabular}{ll}
S \rewr NP VP & (1.000) \\
VP \rewr V NP & (0.733\treesize{$\approx\frac{1+10}{15}$}) \\
VP \rewr V NP PP & (0.267\treesize{$\approx\frac{4}{15}$}) \\  
NP \rewr NP PP & (0.268\treesize{$\approx\frac{1+10}{41}$}) \\
NP \rewr Mary & (0.122\treesize{$\approx\frac{1+4}{41}$}) \\ 
NP \rewr a bird & (0.366\treesize{$\approx\frac{1+4+10}{41}$}) \\
NP \rewr a worm & (0.244\treesize{$\approx\frac{10}{41}$}) \\
PP \rewr on a tree & (1.000) \\  
V \rewr saw & (1.000) \\
\end{tabular}
\end{center}
The probabilities of the rules of this grammar form our first EM
re-estimate $p_1$. So, we are ready for the second EM iteration. The
following table displays the rules of our manually written CFG,
as well as their probabilities allocated by the different EM re-estimates
\begin{center}
\begin{tabular}{l|ccccccccccccccc}
  \textbf{CFG rule}  & $p_0$ & $p_1$ & $p_2$ & $p_3$ & $\dots$ & $p_{18}$
  \\
  \hline
  S \rewr NP VP      & 1.000 & 1.000 & 1.000 & 1.000 &     & 1.000 \\
  VP \rewr V NP      & 0.500 & 0.733 & 0.807 & 0.850 &     & 0.967 \\
  VP \rewr V NP PP   & 0.500 & 0.267 & 0.193 & 0.150 & 	   & 0.033 \\
  NP \rewr NP PP     & 0.250 & 0.268 & 0.287 & 0.298 &     & 0.326 \\
  NP \rewr Mary      & 0.250 & 0.122 & 0.118 & 0.117 &     & 0.112 \\
  NP \rewr a bird    & 0.250 & 0.366 & 0.357 & 0.351 &     & 0.337 \\
  NP \rewr a worm    & 0.250 & 0.244 & 0.238 & 0.234 &     & 0.225 \\
  PP \rewr on a tree & 1.000 & 1.000 & 1.000 & 1.000 &     & 1.000 \\
  V \rewr saw        & 1.000 & 1.000 & 1.000 & 1.000 &     & 1.000 \\
\end{tabular}
\end{center}
In the $19^{th}$ iteration, nothing new happens. So, we can quit the
algorithm and discuss the results. After all, of course, the most
interesting thing for us is how these different PCFGs perform, if they
are applied to ambiguity resolution. So, we will have a look at the
probabilities these PCFGs allocate to the two analyzes of the first
sentence. We have already noted several times that $p$ prefers $x_1$ to
$x_2$, if
\begin{displaymath}
p(\treesize{VP\rewr V NP}) \cdot p(\treesize{NP\rewr NP PP})
> p(\treesize{VP\rewr V NP PP})
\end{displaymath}
The following table displays the values of these terms for our
re-estimated PCFGs
\begin{center}
\begin{tabular}{c|cc}
$p$ & $p(\treesize{VP\rewr V NP}) \cdot p(\treesize{NP\rewr NP PP})$ 
&
$p(\treesize{VP\rewr V NP PP})$
\\
\hline
$p_0$
&
0.500\ $\cdot$\ 0.250\ =\ 0.125
&
0.500 
\\
$p_1$
&
0.733\ $\cdot$\ 0.268\ =\ 0.196
&
0.267
\\
$p_2$
&
0.807\ $\cdot$\ 0.287\ =\ \textbf{0.232}
&
0.193
\\
$p_3$
&
0.850\ $\cdot$\ 0.298\ =\ \textbf{0.253}
&
0.150
\\
$\vdots$
\\
$p_{18}$
&
0.967\ $\cdot$\ 0.326\ =\ \textbf{0.315}
&
0.033
\end{tabular}
\end{center}
So, the EM re-estimates prefer $x_1$ to $x_2$ starting with the second
EM iteration, due to the fact that the term $p(\treesize{VP\rewr V
NP}) \cdot p(\treesize{NP\rewr NP PP})$ is monotonic increasing
(within a range from $0.125$ to $0.315$), while at the same time the
term $p(\treesize{VP\rewr V NP PP})$ is drastically monotonic
decreasing (from $0.500$ to $0.033$).


\section*{Acknowledgments}
Parts of the paper cover parts of the teaching material of two courses
at ESSLLI 2003 in Vienna. One of them has been sponsored by the
\textit{European Chapter of the Association for Computational
Linguistics (EACL)}, and both have been co-lectured by Khalil Sima'an
and me. Various improvements of the paper have been suggested by
Wietse Balkema, Gabriel Infante-Lopez, Karin M\"uller, Mark-Jan Nederhof,
Breannd$\acute{\mbox{a}}$n $\acute{\mbox{O}}$
Nuall$\acute{\mbox{a}}$in, Khalil Sima'an, and Andreas Zollmann.

\bibliographystyle{chicago}

\begin{thebibliography}{}

\bibitem[\protect\citeauthoryear{Backus}{Backus}{1959}]{Backus:1959}
Backus, J.~W. (1959).
\newblock {The syntax and semantics of the proposed international algebraic
  language of the Z\"urich ACM-GAMM Conference}.
\newblock In {\em {Proceedings of the International Conference on Information
  Processing}}, Paris.

\bibitem[\protect\citeauthoryear{Baker}{Baker}{1979}]{Baker:79}
Baker, J.~K. (1979).
\newblock Trainable grammars for speech recognition.
\newblock In D.~Klatt and J.~Wolf (Eds.), {\em Speech Communication Papers for
  {ASA'97}}, pp.\  547--550.

\bibitem[\protect\citeauthoryear{Baum}{Baum}{1972}]{Baum:72}
Baum, L.~E. (1972).
\newblock An inequality and associated maximization technique in statistical
  estimation for probabilistic functions of {M}arkov processes.
\newblock {\em Inequalities\/}~{\em III}, 1--8.

\bibitem[\protect\citeauthoryear{Booth and Thompson}{Booth and
  Thompson}{1973}]{Booth:73}
Booth, T.~L. and R.~A. Thompson (1973).
\newblock Applying probability measures to abstract languages.
\newblock {\em {IEEE} Transactions on Computers\/}~{\em C-22\/}(5), 442--450.

\bibitem[\protect\citeauthoryear{Charniak}{Charniak}{1996}]{Charniak:96}
Charniak, E. (1996).
\newblock Tree-bank grammars.
\newblock Technical Report CS-96-02, Brown University.

\bibitem[\protect\citeauthoryear{Chi}{Chi}{1999}]{Chi:1999}
Chi, Z. (1999).
\newblock Statistical properties of probabilistic context-free grammars.
\newblock {\em Computational Linguistics\/}~{\em 25\/}(1).

\bibitem[\protect\citeauthoryear{Chi and Geman}{Chi and
  Geman}{1998}]{Chi:1998b}
Chi, Z. and S.~Geman (1998).
\newblock Squibs and discussions: Estimation of probabilistic context-free
  grammars.
\newblock {\em Computational Linguistics\/}~{\em 24\/}(2).

\bibitem[\protect\citeauthoryear{Chomsky}{Chomsky}{1956}]{Chomsky:1956}
Chomsky, N. (1956).
\newblock Three models for the description of language.
\newblock {\em IRE Transactions on Information Theory\/}.

\bibitem[\protect\citeauthoryear{Cover and Thomas}{Cover and
  Thomas}{1991}]{CoverThomas:91}
Cover, T.~M. and J.~A. Thomas (1991).
\newblock {\em Elements of Information Theory}.
\newblock New York: Wiley.

\bibitem[\protect\citeauthoryear{DeGroot}{DeGroot}{1989}]{DeGroot:89}
DeGroot, M.~H. (1989).
\newblock {\em Probability and statistics\/} (2 ed.).
\newblock Addison-Wesley.

\bibitem[\protect\citeauthoryear{Dempster, Laird, and Rubin}{Dempster
  et~al.}{1977}]{Dempster:77}
Dempster, A.~P., N.~M. Laird, and D.~B. Rubin (1977).
\newblock Maximum likelihood from incomplete data via the {{\em EM}} algorithm.
\newblock {\em J. Royal Statist. Soc.\/}~{\em 39\/}(B), 1--38.

\bibitem[\protect\citeauthoryear{Duda, Hart, and Stork}{Duda
  et~al.}{2001}]{Duda:01}
Duda, R.~O., P.~E. Hart, and D.~G. Stork (2001).
\newblock {\em Pattern Classification --- 2nd ed}.
\newblock New York: Wiley.

\bibitem[\protect\citeauthoryear{Hopcroft and Ullman}{Hopcroft and
  Ullman}{1979}]{HopcroftUllman:79}
Hopcroft, J.~E. and J.~D. Ullman (1979).
\newblock {\em Introduction to Automata Theory, Languages, and Computation}.
\newblock Reading, MA: Addison-Wesley.

\bibitem[\protect\citeauthoryear{Jaynes}{Jaynes}{1957}]{Jaynes:57}
Jaynes, E.~T. (1957).
\newblock Information theory and statistical mechanics.
\newblock {\em Physical Review\/}~{\em 106}, 620--630.

\bibitem[\protect\citeauthoryear{Johnson}{Johnson}{1998}]{Johnson:1998}
Johnson, M. (1998).
\newblock {PCFG} models of linguistic tree representations.
\newblock {\em Computational Linguistics\/}~{\em 24\/}(4).

\bibitem[\protect\citeauthoryear{Klein and Manning}{Klein and
  Manning}{2003}]{KleinManning:2003}
Klein, D. and C.~D. Manning (2003).
\newblock Accurate unlexicalized parsing.
\newblock In {\em Proceedings of {ACL-03}}, Sapporo, Japan.

\bibitem[\protect\citeauthoryear{Lari and Young}{Lari and
  Young}{1990}]{Lari:90}
Lari, K. and S.~J. Young (1990).
\newblock The estimation of stochastic context-free grammars using the
  inside-outside algorithm.
\newblock {\em Computer Speech and Language\/}~{\em 4}, 35--56.

\bibitem[\protect\citeauthoryear{McLachlan and Krishnan}{McLachlan and
  Krishnan}{1997}]{McLachlan:97}
McLachlan, G.~J. and T.~Krishnan (1997).
\newblock {\em The {EM} Algorithm and Extensions}.
\newblock New York: Wiley.

\bibitem[\protect\citeauthoryear{Nederhof and Satta}{Nederhof and
  Satta}{2003}]{NederhofSatta:2003}
Nederhof, M.-J. and G.~Satta (2003).
\newblock Probabilistic parsing as intersection.
\newblock In {\em {Proceedings of the 8th International Workshop on Parsing
  Technologies (IWPT-03)}}, Nancy, France.

\bibitem[\protect\citeauthoryear{Pereira and Schabes}{Pereira and
  Schabes}{1992}]{Pereira:92}
Pereira, F. and Y.~Schabes (1992).
\newblock Inside-outside reestimation from partially bracketed corpora.
\newblock In {\em Proceedings of {ACL'92}}, Newark, Delaware.

\bibitem[\protect\citeauthoryear{Prescher}{Prescher}{2001}]{Prescher:01c}
Prescher, D. (2001).
\newblock Inside-outside estimation meets dynamic {EM}.
\newblock In {\em Proceedings of IWPT-2001}, Beijing.

\bibitem[\protect\citeauthoryear{Prescher}{Prescher}{2002}]{Prescher:Diss}
Prescher, D. (2002).
\newblock {\em {EM-basierte maschinelle Lernverfahren f\"ur nat\"urliche
  Sprachen}}.
\newblock Ph.\ D. thesis, IMS, University of Stuttgart.

\bibitem[\protect\citeauthoryear{Ratnaparkhi}{Ratnaparkhi}{1997}]{Rat:97Report}
Ratnaparkhi, A. (1997).
\newblock A simple introduction to maximum-entropy models for natural language
  processing.
\newblock Technical report, University of Pennsylvania.

\bibitem[\protect\citeauthoryear{Sanchez and Benedi}{Sanchez and
  Benedi}{1997}]{Sanchez1997}
Sanchez, J.~A. and J.~M. Benedi (1997).
\newblock Consistency of stochastic context-free grammars from probabilistic
  estimation based on growth transformations.
\newblock {\em IEEE Transactions on Pattern Analysis and Machine
  Intelligence\/}~{\em 19}.

\bibitem[\protect\citeauthoryear{Wu}{Wu}{1983}]{Wu:83}
Wu, C. F.~J. (1983).
\newblock On the convergence properties of the {EM} algorithm.
\newblock {\em The Annals of Statistics\/}~{\em 11\/}(1), 95--103.

\end{thebibliography}

\end{document}